\documentclass[12pt,a4paper,twoside]{article}
\usepackage[english]{babel} 
\usepackage{csquotes}
\usepackage[T1]{fontenc} 
\usepackage[cyr]{aeguill} 
\usepackage{epsfig} 
\usepackage{amsmath,amsthm, mathtools,thmtools} 
\usepackage{amsfonts,amssymb,bm}
\usepackage{algorithm2e} 
\RestyleAlgo{ruled}
\usepackage{url} 
\usepackage{xcolor}
\usepackage{hyperref}  

\usepackage{cleveref} 
\crefname{figure}{figure}{figures}
\Crefname{figure}{Figure}{Figures}
\crefname{property}{property}{properties}
\Crefname{property}{Property}{Properties}
\Crefformat{appendix}{Section #2#1#3 of Supplementary Material}
\usepackage[a4paper,margin=0.75in]{geometry}
\usepackage[super]{nth}
\usepackage{float} 
\usepackage{caption} 
\DeclareCaptionLabelSeparator{hyphen}{\space--\space}
\usepackage[mode=buildnew]{standalone}
\usepackage{multirow}

\definecolor{blueind}{RGB}{016, 101, 171}
\definecolor{cyanind}{RGB}{058, 147, 195}
\definecolor{electricblue}{RGB}{142, 196, 222}
\definecolor{greenind}{RGB}{112, 130, 56}
\definecolor{burntorange}{RGB}{179, 021, 041}
\definecolor{goldenyellow}{RGB}{215, 095, 076}
\definecolor{peach}{RGB}{246, 164, 130}

\definecolor{bluefonceps}{RGB}{0,78,125}
\definecolor{blueps}{RGB}{14,135,201}
\definecolor{vertps}{RGB}{64,183,105}

\usepackage[style=apa, sorting=nyt, backend=biber]{biblatex}
\AtEveryCite{\color{blue}}
\AtEveryBibitem{%
  \clearfield{annotation} 
}

\usepackage{changepage}
\usepackage{sectsty}

\newcommand{\eps}[1][]{\ensuremath{\epsilon_{#1}}}

\newcommand{\supp}[1]{S^{(#1)}}
\newcommand{\bothsupps}{\supp{1,2}} 
\newcommand{\E}{\mathbb{E}}

\newcommand{\vbound}{\mathcal{J}}
\newcommand{\Rest}{\widehat{\mathcal{R}}}

\newcommand{\Entropy}{\mathcal{H}}
\newcommand{\bX}{\bm{X}}
\newcommand{\bZ}{\bm{Z}}
\newcommand{\bW}{\bm{W}}
\newcommand{\di}{\mathrm{d}}
\newcommand{\thetasupp}{\bm{\theta}_{\bothsupps}}
\newcommand{\alphasupp}{\bm{\alpha}_{\bothsupps}}
\newcommand{\pisupp}{\bm{\pi}_{\supp{1}}}
\newcommand{\rhosupp}{\bm{\rho}_{\supp{2}}}

\newcommand{\admissupp}[1]{\mathcal{S}_{Q_{#1}}}

\newcommand{\pen}{\text{pen}}

\newcommand{\pmis}{p_{\text{mis}}}

\DeclareMathAlphabet{\mathbbb}{U}{bbold}{m}{n} 
\DeclareMathOperator*{\argmax}{arg\,max}
\DeclareMathOperator*{\BICL}{BIC-L}
\DeclareMathOperator*{\ICL}{ICL}

\newtheorem{property}{Property}

\usepackage{enumitem}

\usepackage{authblk}
\title{Common Structure Discovery in Collections of Bipartite Networks: Application to Pollination Systems}

\date{}

\author[1]{Louis Lacoste}
\author[1]{Pierre Barbillon}
\author[1]{Sophie Donnet}
\affil[1]{Université Paris-Saclay, AgroParisTech, INRAE,
UMR MIA Paris-Saclay, 91120, Palaiseau, France}

\usepackage{ifthen}
\newif\ifSUPPONLY
\ifthenelse{\equal{\detokenize{supplementaries}}{\jobname}}{
    \SUPPONLYtrue 
}{
    \SUPPONLYfalse 
}
\newif\ifNOSUPP
\ifthenelse{\equal{\detokenize{article-only}}{\jobname}}{
    \NOSUPPtrue 
}{
    \NOSUPPfalse 
}

\ifNOSUPP
\else
\usepackage{pgffor}
\fi

\begin{document}

\ifSUPPONLY
\else
    \maketitle

\begin{abstract}
  Bipartite networks are widely used to encode the ecological interactions. Being able to compare the organization of  bipartite networks is a first step toward a better understanding of  how environmental factors shape community structure and resilience.
  Yet current methods for structure detection in bipartite networks overlook shared patterns across collections of networks.
  We introduce the \emph{colBiSBM}, a family of probabilistic models for
  collections of bipartite networks that extends the classical Latent Block Model (LBM). The proposed framework assumes that networks are independent
  realizations of a shared mesoscale structure, encoded through common
  inter-block connectivity parameters.
  We establish identifiability conditions for the different variants of
  \emph{colBiSBM} and develop a variational EM algorithm for parameter
  estimation, coupled with an adaptation of the integrated classification
  likelihood (ICL) criterion for model selection. We demonstrate how our approach can be used to classify networks based on their topology or organization.
  Simulation
  studies highlight the ability of \emph{colBiSBM} to recover common
  structures, improve clustering performance, and enhance link prediction by
  borrowing strength across networks.
  An application to plant--pollinator networks highlights how the method uncovers
  shared ecological roles and partitions networks into sub-collections with
  similar connectivity patterns. These results illustrate the methodological and
  practical advantages of joint modeling over separate network analyses in the
  study of bipartite systems.
\end{abstract}

\section{Introduction}
\label{sec:introduction}
\paragraph{Context}
Plant--pollinator networks are increasingly documented across the globe
(see~\cite{doreRelativeEffectsAnthropogenic2021} and references therein). These
systems belong to the broader class of bipartite graphs, which also capture
other types of ecological interactions between two sets of actors, such as
bird--seed dispersal, prey--predator, or host--parasite relationships
(\cite{kaszewska-gilasGlobalStudiesHostParasite2021,WebLifeEcological2022}).

Bipartite
graphs are widely applied in biology, not only for ecological networks but also
in fields such as medicine, where they are used to model biomedical,
biomolecular, and epidemiological networks
(\cite{pavlopoulosBipartiteGraphsSystems2018}). Numerous metrics are commonly
employed to characterize bipartite networks, including connectivity and
nestedness (see~\cite{corsoConnectivityNestednessBipartite2011}). However,
methods for comparing bipartite networks based on their topology remain under
active development.
Advancing our understanding of bipartite networks and the functional roles of
communities is crucial for ecosystem conservation and management
(\cite{sheykhaliRobustnessExtinctionPlasticity2020,elleUsePollinationNetworks2012}).

A flexible and widely used approach for uncovering mesoscale structure in
bipartite networks is the Latent Block Model (LBM)
(\cite{govaertClusteringBlockMixture2003,govaertLatentBlockModel2010}) also known as biSBM (for bipartite SBM). In this
model, each node is associated with a latent variable that assigns it to a
functional group or block. Nodes in the same block are assumed to have similar
connectivity patterns, so biSBM grouping can reveal functional organization, with
potential ecological interpretation. However, the standard biSBM applies to a
single network at a time. While one could independently infer biSBMs for each
network in a collection and then compare their structures, this approach has
clear limitations: in particular, block assignments are not guaranteed to
correspond across networks. Being able to compare bipartite networks in ecology
would provide an interesting framework to uncover how species interactions
shape biodiversity, ecosystem stability, and functional resilience, offering
crucial insights for understanding environmental change and guiding
conservation efforts.
This motivates us to develop a new model for structure detection in collections
of bipartite networks.

\paragraph{Our contribution}
We propose to jointly model a collection of bipartite networks by extending the
LBM. We assume that the networks are independent realizations of the biSBM,
sharing common parameters. From this assumption, an interesting implication is
that the groups represent a shared connectivity structure, which can be
interpreted as functional groups. Note that we consider networks involving
their own nodes, no assumption of correspondence between nodes across networks
is made. If some networks involve common individuals (species) then, our
approach does not take this information into account and, possibly, the nodes
representing the same individual/species across multiple networks may be
allocated to different blocks.
The proposed model is referred to as the \emph{colBiSBM}, standing for
\emph{collection of bipartite stochastic block models}. We introduce several
variants of the model, including one where the networks are assumed to be
identically distributed according to the same biSBM. Recognizing that this
assumption may be overly restrictive for real-world data, we propose an
alternative that allows for fluctuations in block proportions, including the
possibility of certain blocks being unpopulated. This flexibility enables
networks to share only parts of their connectivity structures.

Inference of the parameters, block memberships, and model selection is
performed using established tools for SBM and LBM models, building on the work
of~\cite{chabert-liddellLearningCommonStructures2024}. Specifically, we employ
Variational Expectation-Maximization and an adaptation of the Integrated
Classification Likelihood (ICL,~\cite{daudinMixtureModelRandom2008}) for model
selection, incorporating novel strategies for model-space exploration.

Three key features of the \emph{colBiSBM} model are particularly noteworthy.
First, it enables the identification of a shared connectivity structure that
accounts for the observed collection of networks, while also assessing whether
this joint modeling approach is appropriate for the collection. By sharing
connectivity and, consequently, functional groups, the model facilitates the
identification of nodes that exhibit similar ecological roles across different
networks. Second, we show how our model can be used to partition the collection of networks
thus leading to sub-collections of networks that
are  similar in terms of their connectivity structures. These two
capabilities address critical needs in ecological research, as highlighted by
~\cite{michalska-smithTellingEcologicalNetworks2019,pichonTellingMutualisticAntagonistic2024}.
Finally, a notable advantage of joint modeling is its ability to transfer
information between networks that share a common structure. This approach
improves the recovery of more detailed connectivity structures and block
memberships, while also facilitating the prediction of missing
links~(\cite{clausetHierarchicalStructurePrediction2008}) by utilizing
connectivity patterns observed throughout the entire collection, thus using
mesoscale information instead of local
information~(\cite{zhouPredictingMissingLinks2009}) to predict missing links.

\paragraph{Related work}
Ecological studies increasingly analyze collections of networks rather than
single systems. For example,~\cite{doreRelativeEffectsAnthropogenic2021}
investigate anthropogenic pressures across a large dataset of plant--pollinator
networks, examining species richness, connectance, and the influence of
covariates. Similarly,~\cite{pichonTellingMutualisticAntagonistic2024} use
multi-scale structural descriptors--such as degree distributions, motif
frequencies, and nestedness--to differentiate mutualistic and antagonistic
networks, identifying structural signatures of interaction types.
For pairwise graph comparison, one can rely on dedicated metrics
(see~\cite{willsMetricsGraphComparison2020a}). More recently, graph embedding
techniques (see review by~\cite{xuUnderstandingGraphEmbedding2021}), often based on
neural networks, have become popular: they map graphs into a low-dimensional
latent space, making clustering-based comparisons possible. While these
approaches are highly scalable, they generally lack the interpretability of
block model based methods.
Joint modeling of network collections has been studied mostly in the context of
unipartite networks. For instance,
\cite{arroyoInferenceMultipleHeterogeneous2021} introduce the \emph{common
    subspace independent edge} random graph model, which captures a shared
invariant subspace across networks. Building on SBM theory,
\cite{agterbergJointSpectralClustering2025} analyse the multilayer
degree-corrected stochastic block model (ML-DCSBM), proposing a spectral
algorithm that aggregates normalized embeddings across layers. Both approaches
assume that nodes are shared across networks in a multilayer collection. In
contrast,~\cite{kumpulainenYourBlockOur2024} consider collections of SBMs that
share a \emph{core} subset of $s$ blocks, with the remaining blocks being
network-specific. Finally,~\cite{matiasStatisticalClusteringTemporal2017}
address temporal network sequences, where the underlying connectivity structure
evolves over time.

\paragraph{Outline}
The remainder of this paper is organized as follows.
In~\Cref{ssec:data-motivation-and-bipartite-stochastic-block-model}, we
introduce the biSBM  for a single bipartite network and apply it
independently to a collection of four plant-pollinator networks.
\Cref{ssec:joint-modelisation-of-multiple-bipartite-networks} presents the \emph{colBiSBM} model and its variants.  Subsequently,~\Cref{ssec:likelihood-of-the-models,ssec:identifiability-of-the-models}
provide  the likelihood of the models and establish the identifiability of their
parameters.
In~\Cref{ssec:variational-estimation-of-the-parameters,ssec:model-selection},
we detail the inference strategy (parameter estimation, selection of the numbers of blocks).
In~\Cref{sssec:testing-common-connectivity-structure}, we demonstrate that our method enables the partitioning of a collection of networks into sub-collections sharing a common mesoscopic organization.
Finally,
in~\Cref{sec:applications-on-ecological-networks}, we apply our methodology to a collection of five  plant-pollinator networks and stress its benefits.

\section{Data motivation and existing models}
\label{sec:data-motivation-and-bipartite-stochastic-block-model}
\subsection{Plant-pollinator networks in British cities}
\label{ssec:data-motivation-and-bipartite-stochastic-block-model}
Let us consider a
collection of $M$ bipartite networks, each of them  representing an (eco)-system involving its own entities.  Each bipartite network $m$ has two sets of nodes of respective sizes  $n_1^m$ and  $n_2^m$, and the interactions only occur between nodes of set $1$ and   of set $2$.
For each network, one can define its bi-adjacency matrix  $(X^m)_{m\in\{1,\dots,M\}}$ of  size  $n_1^m \times n_2^m$ such that  $\forall (i,j) \in[\![1,n_1^m]\!] \times [\![1,n_2^m]\!]$:
\begin{align*}
    \begin{cases}
        X_{ij}^m = 0    & \text{if no interaction is observed between species }i\text{ and }j \text{ in network }m \\
        X_{ij}^m \neq 0 & \text{otherwise.}
    \end{cases}
\end{align*}
With this formalism, the nodes of set $1$  (e.g.,  pollinators)  are in row and the nodes of set $2$  (e.g.,  plants)  are in column. In the following we will talk about row and column nodes.
If the network represents binary observations (like presence-absence) then
$X^m_{ij}\in \{0,1\}$; if the interactions are weighted (like an abundance
count), $X^m_{ij}\in \mathbb{N}$. In what follows, we assume that all the
networks in the collection have the same valuation of the interactions. We
denote by $\bm{X} = (X^1, \dots, X^M)$ the collection of bi-adjacency matrices.

\paragraph{Plant-pollinator networks in British cities}
We present here a collection of $M=4$ plant-pollinator networks
from~\cite{baldockSystemsApproachReveals2019}. This paper studies the
diversity, robustness and impact of the type of environment on the ecological
aspect of plant-pollinator networks in four major British cities, namely
Bristol, Edinburgh, Leeds and Reading. The original interactions are abundance
counts but, in order to remove potential sampling or species abundance effects,
we consider them as binary interactions by setting $X_{ij}^m = 1$ if pollinator
$i$ has visited plant $j$ at least once. Their characteristics in terms of
number of nodes and density are provided in \Cref{tab:baldock-charateristics}.
Note that at most 35\% of the species (plants or pollinators) are shared by any
pair of networks. Additional details on the networks are provided
in\ifNOSUPP~supplementaries.\else~\Cref{tab:baldock-shared-species}\fi.


\begin{table}
    \centering
    \caption{Characterics of the four  British plant-pollinator networks \label{tab:baldock-charateristics}}
    \begin{tabular}{cccccc}
        \hline
        $m$ & Place     & nb. of pollinators & nb. of plants & nb. of          & density \\
            &           & ($n_1^m$)          & ($n_2^m$)     & $\neq 0$ inter. &         \\
        \hline
        1   & Bristol   & 180                & 193           & 662             & 0.019   \\
        2   & Edinburgh & 129                & 137           & 425             & 0.024   \\
        3   & Leeds     & 126                & 154           & 573             & 0.029   \\
        4   & Reading   & 167                & 156           & 574             & 0.022   \\
        \hline
    \end{tabular}
\end{table}

We are interested in the global organization of those four networks and not
specifically on the role played by each species in its ecosystem. When willing
to understand how the networks are organized, the Stochastic Block Model (in
short, SBM) has proved its capacity to capture various complex structures,
(see~\cite{sanderWhatCanInteraction2015} for instance). This model has been
extended to bipartite networks resulting into the bipartite Stochastic Block
Model (biSBM), also known as the Latent Block
Model~(\cite{govaertClusteringBlockMixture2003}). 

\subsection{Separate biSBM (sep-biSBM)}\label{ssec:separate-bisbm-sep-bisbm}

The first naive way to deal with multiple networks is to adjust the biSBM
separately on each network and then "compare" the resulting structures. The
biSBM is a latent variable model, assuming that the row and column nodes of
each network $m$ are divided into respectively $Q_1^{m}$ and $Q_2^{m}$ blocks.
Let $Z^m=(Z^m_1, \dots, Z^m_i, \dots, Z^m_{n_1^m})$ and $W^m = (W^m_1,
    \dots,W^m_j, \dots, W^m_{n_2^m})$ encode the belongings of the nodes to their
blocks: $Z^m_i = q$ if row node $i$ of network $m$ belongs to row block $q$
($q\in\{1,\dots,Q_1^{m}\}$) and $W^m_j = r$ if column node $j$ of network $m$
belongs to column block $r$ ($r\in\{1,\dots,Q_2^{m}\}$). The $Z_i^m$ and
$W_j^m$ are independent categorical variables with probability:
\begin{align}\label{eqn:lbm-block-membership-prob}
    \mathbb{P}(Z_i^m=q)=\pi_q^m, &  & \mathbb{P}(W_j^m=r)=\rho_r^m
\end{align}
where $\pi_q^m > 0$, $\rho_r^m > 0$, $\sum_{q=1}^{Q_1^{m}}\pi_q^m = 1$ and
$\sum_{r=1}^{Q_2^{m}}\rho_r^m = 1$. Given the latent variables
$(Z^m, W^m)$, the $X_{ij}^m$'s are assumed to be independent and distributed
as
\begin{align}\label{eqn:lbm-conditional-to-latent}
    X_{ij}^m \mid Z_i^m = q,W_j^m = r \sim \mathcal{F}(.;\alpha_{qr}^m)
\end{align}
where $\mathcal{F}$ is referred to as the emission distribution. $\mathcal{F}$ is chosen to
be the Bernoulli distribution for binary interactions, and the Poisson
distribution for weighted interactions such as counts. Let $f$ be the density of
the emission distribution, then:
\begin{equation}\label{eqn:lbm-emission}
    \log f(X^m_{ij};\alpha_{qr}^m) =
    \begin{cases}
        X_{ij}^m \log(\alpha_{qr}^m) + (1-X_{ij}^m) \log(1-\alpha_{qr}^m) & \text{for Bernoulli emission} \\
        -\alpha_{qr}^m + X_{ij}^m \log(\alpha_{qr}^m) - \log(X_{ij}^m!)   & \text{for Poisson emission}
    \end{cases}
\end{equation}
Equations~\eqref{eqn:lbm-block-membership-prob},
\eqref{eqn:lbm-conditional-to-latent} and~\eqref{eqn:lbm-emission} define the
biSBM model, and we will now use a short notation:
\begin{align}
    \tag{\emph{sep-biSBM}}
    X^m \sim \mathcal{F}\text{-biSBM}_{n_1^m,n_2^m}(Q_1^{m}, Q_2^{m}, \bm{\pi}^m, \bm{\rho}^m, \bm{\alpha}^m) &  & \forall m = 1, \dots M
\end{align}
where $\mathcal{F}$ encodes the emission distribution, $n_1^m,n_2^m$ are the numbers of row
and column nodes, $Q_1^{m}, Q_2^{m}$ are the numbers of row and column blocks in
network $m$, $\bm{\pi}^m~=~{(\pi^m_q)}_{q=1,\dots,Q_1^{m}}$ and
$\bm{\rho}^m~=~{(\rho^m_r)}_{r=1,\dots,Q_2^{m}}$ are the block proportion  vectors. The $Q_1^{m} \times Q_2^{m}$ matrix
$\bm{\alpha}^m = {(\alpha^m_{qr})}_{q = 1,\dots,Q_1^{m},r = 1,\dots,Q_2^{m}}$
contains the connectivity parameters, i.e.,~the parameters of the emission distribution.
In this sep-biSBM model, we suppose that each network $m$ follows a biSBM with
its own parameters $\bm{\theta}^m = (\bm{\pi}^m, \bm{\rho}^m, \bm{\alpha}^m)$.
Each model is defined up to label switching of the blocks, as the order of the
blocks is completely arbitrary. \\

For given numbers of blocks $(Q_1^{m}, Q_2^{m})$, the parameters
$\bm{\theta}^m$ can be estimated by a variational version of the EM algorithm.
This method is implemented in the \texttt{R} package
\textsf{sbm}~(\cite{chiquetSbmStochasticBlockmodels2024,
    legerBlockmodelsLatentStochastic2021}). The number of blocks $Q_1^{m}$ and
$Q_2^{m}$ can be chosen through the Integrated Classification Likelihood
criterion (ICL). Details on the variational EM and ICL will be provided in the
following section. The biSBM is an unsupervised clustering method in that the
blocks cluster the nodes that \enquote{behave the same} in the network. The
nodes are assigned to blocks by their maximum posterior probability:
\[
    \begin{array}{ccl}
        \widehat{Z}^m_i & = & \argmax_{q=1, \dots Q_1^{m}} \mathbb{P}(Z^m_i=q \mid X^m, \widehat{\theta}^m ), \\
        \widehat{W}^m_j & = & \argmax_{r=1, \dots Q_2^{m}} \mathbb{P}(W^m_j=r \mid X^m,\widehat{\theta}^m).
    \end{array}
\]

\paragraph{Application on the plant-pollinator networks}

We applied the sep-biSBM model on the four British plant-pollinator networks
refereed as Bristol, Edinburgh, Leeds and Reading. 
We obtain $4$ blocks of pollinators for all the networks. The number of blocks
of plants is chosen to be $3$ in the Bristol, Edinburgh and Reading networks
but only $2$ in the Leeds network. In Figure~\ref{fig:lbm-plant-pollinator},
the four bi-adjacency matrices are arranged according to the blocks obtained by
the inference procedure. For each pair of blocks $(q,r)$, the background color
of the submatrix is chosen to represent the intensity of the interaction
$\alpha^m_{qr}$. As previously mentioned, the labels assigned to the blocks are
entirely arbitrary. Since we aim at comparing the estimated organisation of the
networks, we chose to order the row and column blocks with respect to their
marginal density. This choice underscores the nestedness present within the
networks (a structural element that is frequently observed in plant-pollinator
networks, see e.g.,~\cite{corsoConnectivityNestednessBipartite2011}). A quick
look at \Cref{fig:lbm-plant-pollinator} suggests the existence of structural
similarities in the four organisations and indicates that a joint modeling
approach could be beneficial. In the following, we propose several
probabilistic models that assume a connectivity structure shared by all the
networks of interest.

\begin{figure}[t]
    \centering
    \includegraphics[width=0.9\linewidth]{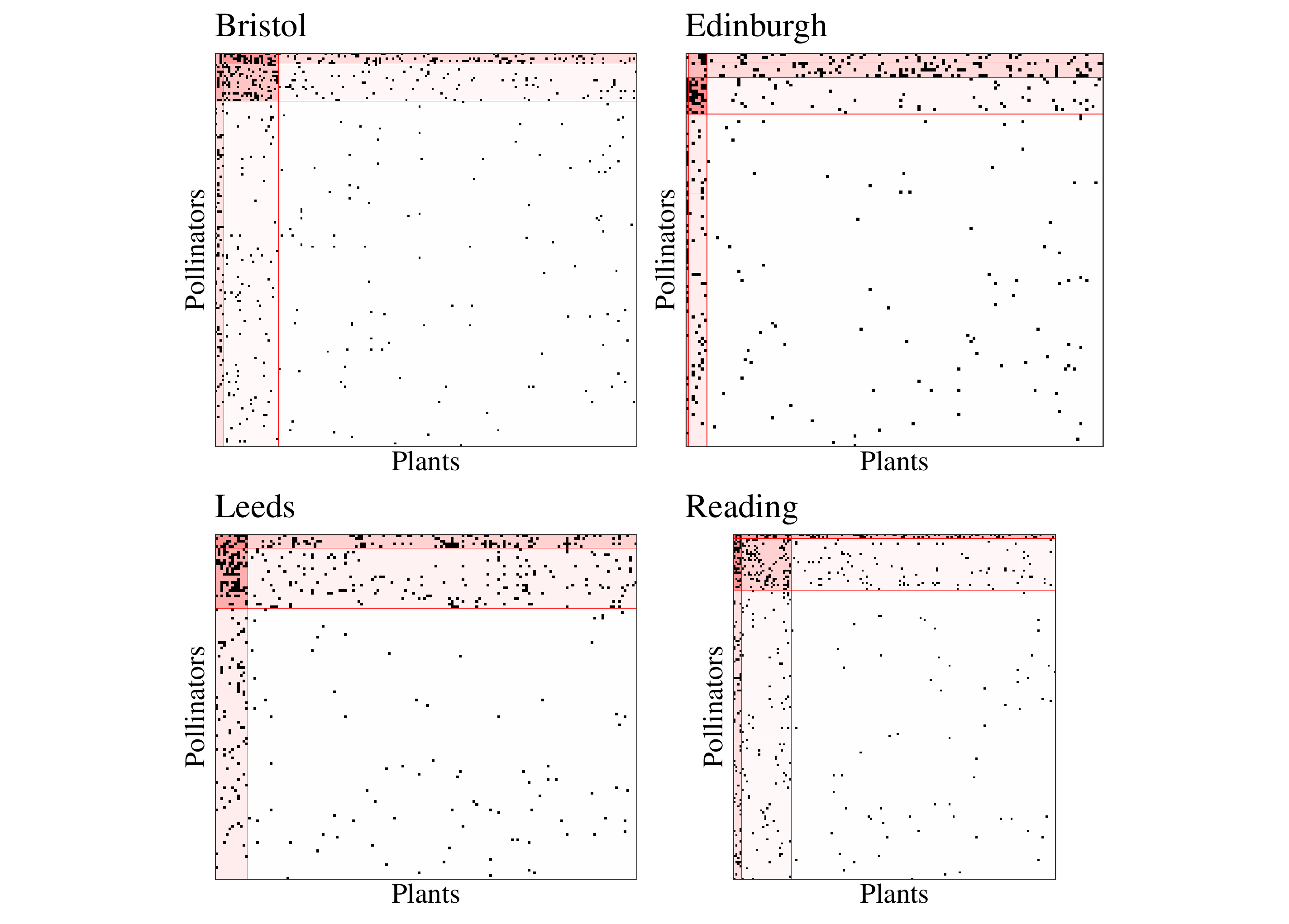}
    \caption{Matrix representation of the 4 British plant-pollinator networks. The blocks
        were obtained by fitting a biSBM on each network separately.The blocks were reordered following the marginal probability of connection.  The shades of red show the connectivity parameters $\widehat{\alpha}_{qr}^m$.  }\label{fig:lbm-plant-pollinator}
\end{figure}


\section{Joint modelisation of multiple bipartite networks }  
\label{ssec:joint-modelisation-of-multiple-bipartite-networks}

We propose several models adapted to collections of networks. All of them
assume that the $(X^m)_{m=1,\dots,M}$ are independent realizations of a biSBM
model.

\subsection{A collection of i.i.d. biSBM}\label{ssec:a-collection-of-i-i-d-bipartite-sbm}
This first model assumes that all the networks are independent realizations of
the same $Q_1$-$Q_2$-biSBM with identical parameters. The iid-colBiSBM is
defined as follows:

\begin{align}
    \tag{iid-colBiSBM}
    X^m \sim \mathcal{F}\text{-biSBM}_{n_1^m,n_2^m} (Q_1, Q_2, \bm{\pi}, \bm{\rho}, \bm{\alpha}), &  & \forall m = 1, \dots M
\end{align}
where $\forall (q,r)$ 
$\pi_q >0, \rho_r >0$, $\sum_{q=1}^{Q_1} \pi_q  =  \sum_{r=1}^{Q_2} \rho_r = 1 $.
This model involves
\begin{equation*}
    \text{NP}(\text{iid-colBiSBM}) = (Q_1 - 1) + (Q_2 - 1) + Q_1\times Q_2,
\end{equation*}
parameters, the two
first terms corresponding to the row and column block proportions
and the third term to connectivity parameters.

By considering the same biSBM model for all the networks, we assume that the
systems they represent have a similar global functioning. More interestingly,
the inferred blocks in each network will gather species that play the same role
across the networks. The role played by each block/cluster in the system is
encoded in the unique connectivity matrix $\bm{\alpha}$. However, the
iid-colBiSBM also assumes that the block proportions are the same across
networks, which means that each block is represented in the same proportion in
each network. This is a strong assumption which may not be realistic in
plant-pollinator networks and is relaxed in the next section.

\subsection{A collection of bipartite SBM with varying block size on rows or columns}\label{ssec:a-collection-of-bipartite-sbm-with-varying-block-size-on-rows-or-columns}

\paragraph{$\pi\rho$-colBiSBM model}

Our $\pi\rho$-colBiSBM model assumes that the networks share a common
connectivity structure represented by $\bm{\alpha}$ but that each network has
its own row and column block proportions. The $X^m$ are assumed to be
independent and distributed as
\begin{align}
    \tag{\emph{$\pi\rho$}-colBiSBM}
    X^m \sim \mathcal{F}\text{-biSBM}_{n_1^m,n_2^m} (Q_1, Q_2, \bm{\pi}^m, \bm{\rho}^m, \bm{\alpha}), &  & \forall m = 1, \dots M
\end{align}
where $\forall (q,r)$,
$\pi^m_q \geq 0$, $\rho^m_r\geq 0$ and  $\sum_{q=1}^{Q_1} \pi^m_q= \sum_{r=1}^{Q_2} \rho^m_r ~=~1$.

This model is more flexible than the iid-colBiSBM as it allows the row block
and column block proportions to vary between networks. In addition, we allow
some of the $(\pi^m_q)$'s and $(\rho^m_r)$'s to be null ($\pi^m_q\in\left[ 0,1
        \right]$): if $\pi_q^m = 0$ or $\rho_r^m=0$ then the row block $q$ or column
block $r$ is not represented in the network $m$. The connectivity structure of
network $m$ may thus be a subset of a large connectivity structure common to
all networks. The possible nullity of some block proportion raises
identifiability issues which can be tackled if we assume that each block q is
represented in at least one network, i.e., \begin{align*}
    \forall q \in \{1,\dots,Q_1\}, \exists m\in\{1,\dots,M\} & \mbox{  such that }   \pi^m_q > 0  \\
    \forall r \in \{1,\dots,Q_2\}, \exists m\in\{1,\dots,M\} & \mbox{  such that }   \rho^m_r > 0
\end{align*}
Let
us define the support matrix for row  blocks $S^{(1)}$ of size $M\times Q_1$
such that $\forall (m,q),~S^{(1)}_{mq} = \mathbbb{1}_{\pi^m_q > 0}$. Then the
set of admissible supports is the set of binary matrices with a least one
non-null element in each row and column:
\[
    \admissupp{1} \coloneq \biggl\{S^{(1)} \in \mathcal{M}_{M\times Q_1}(\{0,1\})~|~\forall m,~\sum_{q=1}^{Q_1} S^{(1)}_{mq} > 0,~\forall q,~\sum_{m=1}^{M} S^{(1)}_{mq} > 0\biggr\}.
\]
We define equivalently the support matrix for column blocks $S^{(2)}$
For given numbers of blocks $(Q_1,Q_2)$ and support matrices $S^{(1)}$,
$S^{(2)}$, the number of parameters is:
\begin{equation*}
    \text{NP}(\pi\rho\text{-colBiSBM}) = \sum_{m=1}^{M}\Bigg( \sum_{q=1}^{Q_1} S^{(1)}_{mq} - 1 \Bigg) + \sum_{m=1}^{M}\Bigg( \sum_{r=1}^{Q_2} S^{(2)}_{mr} - 1 \Bigg) + \sum_{q,r=1}^{Q_1,Q_2} \mathbbb{1}_{{(S^{(1)\top}S^{(2)})}_{qr}>0}~,
\end{equation*}
where $S^{(1)\top}$ is the transpose of $S^{(1)}$.
The third quantity takes into account that with $\pi\rho$-colBiSBM some row and
column blocks may never interact and thus the corresponding $\alpha_{qr}$ is
not required to define the model.

For the $\pi\rho$-colBiSBM model, we introduce the following notations which
will be useful in the next sections.
\begin{equation}\label{eq:dimS1S2}
    \begin{array}{cccccc}
        \mathcal{Q}_1^m & = & \{q\in\{1,\dots,Q_1\} | \pi_q^m > 0\},     & Q_1^{m} & = & |\mathcal{Q}_1^m| \\
        \mathcal{Q}_2^m & = & \{r \in \{1,\dots,Q_2\} | \rho_r^m > 0\} , & Q_2^{m} & = & |\mathcal{Q}_2^m|
    \end{array}
\end{equation}
$Q_1^{m}$ and $Q_2^{m}$ are the numbers of non-empty row and column blocks in network $m$.


\paragraph{$\pi$-colBiSBM and $\rho$-colBiSBM models}

Some intermediate models can be naturally defined. The $\pi$-colBiSBM allows
the row proportions to vary across networks, while keeping the column
proportions identical for all networks. The $\rho$-colBiSBM is defined
analogously by interchanging rows and columns.

All the models previously defined are summarized in Table
\ref{tab:model-summary}. 

\begin{table}[h]
    \centering
    \caption{Summary of the different models defined in~\Cref{ssec:joint-modelisation-of-multiple-bipartite-networks}. The last line
        is for separate modelisation as presented in~\Cref{sec:data-motivation-and-bipartite-stochastic-block-model}.\label{tab:model-summary}}
    \begin{tabular}{|c|c|c|c|c|}
        \hline
        \hline
        \multirow{2}{*}{Model name}         & Row                                           & Column                                           & Connection                                     & Number                             \\
                                            & Block prop.                                   & Block prop.                                      & param.                                         & of param.                          \\
        \hline
        \multirow{2}{*}{iid-colBiSBM}       & \multirow{2}{*}{$\pi_q^m = \pi_q,~\pi_q > 0$} & \multirow{2}{*}{$\rho_r^m = \rho_r,~\rho_r > 0$} & \multirow{2}{*}{$\alpha^m_{qr} = \alpha_{qr}$} & $(Q_1 - 1) + $                     \\
                                            &                                               &                                                  &                                                & $(Q_2 - 1) + Q_1 Q_2$              \\
        \hline
        \multirow{2}{*}{$\pi$-colBiSBM}     & \multirow{2}{*}{$\pi_q^m \geq 0$}             & \multirow{2}{*}{$\rho_r^m = \rho_r,~\rho_r > 0$} & \multirow{2}{*}{$\alpha^m_{qr} = \alpha_{qr}$} & $\leq M(Q_1 - 1) +$                \\
                                            &                                               &                                                  &                                                & $(Q_2 - 1) + Q_1 Q_2$              \\
        \hline
        \multirow{2}{*}{$\rho$-colBiSBM}    & \multirow{2}{*}{$\pi_q^m = \pi_q,~\pi_q > 0$} & \multirow{2}{*}{$\rho_r^m \geq 0$}               & \multirow{2}{*}{$\alpha^m_{qr} = \alpha_{qr}$} & $\leq (Q_1 - 1) + $                \\
                                            &                                               &                                                  &                                                & $M(Q_2 - 1) + Q_1 Q_2$             \\
        \hline
        \multirow{2}{*}{$\pi\rho$-colBiSBM} & \multirow{2}{*}{$\pi_q^m \geq 0$}             & \multirow{2}{*}{$\rho_r^m \geq 0$}               & \multirow{2}{*}{$\alpha^m_{qr} = \alpha_{qr}$} & $\leq M(Q_1 - 1) + $               \\
                                            &                                               &                                                  &                                                & $M(Q_2 - 1) + Q_1 Q_2$             \\
        \hline
        \multirow{2}{*}{sep-biSBM}          & \multirow{2}{*}{$\pi_q^m > 0$}                & \multirow{2}{*}{$\rho_r^m > 0$}                  & \multirow{2}{*}{$\alpha^m_{qr}$}               & $\sum_{m=1}^{M} [(Q_1^{m} - 1) + $ \\
                                            &                                               &                                                  &                                                & $(Q_2^{m} - 1) + Q_1^{m} Q_2^{m}]$ \\
        \hline
    \end{tabular}
\end{table}

\section{Statistical inference }
\label{sec:inference-identifiability-and-model-selection}

We now address the task of statistical inference. More precisely, we derive the
expression of the likelihood, establish the identifiability of the parameters,
propose a variational version of the EM algorithm to estimate the parameters,
and finally suggest a penalized likelihood criterion to infer the number of
blocks. The results and methods are given for $\pi\rho$-colBiSBM. The
corresponding formula for $\pi$-colBiSBM and $\rho$-colBiSBM are left to the
reader.

\subsection{Log-likelihood expression for $\pi\rho$-colBiSBM}
\label{ssec:log-likelihood-expression-for-pi-rho-colbisbm}


\label{ssec:likelihood-of-the-models}
For the given matrices $S^{(1)}, S^{(2)}$, let $\bm{\theta}_{S^{(1)},S^{(2)}}$ be:
\[
    \bm{\theta}_{S^{(1)},S^{(2)}} = \left(\bm{\pi}^1, \dots ,\bm{\pi}^M, \bm{\rho}^1,
    \dots, \bm{\rho}^M, \bm{\alpha}\right) = \left(\bm{\pi}, \bm{\rho}, \bm{\alpha}\right)
\]
where $\pi_q^m = 0$ for any $q$ such that $S^{(1)}_{mq} = 0$ and $\rho_r^m = 0$
for any $r$ such that $S^{(2)}_{mr} = 0$. $\pi\rho$-colBiSBM is a latent
variable model and consequently, the observed log-likelihood
$\ell(\bm{X};\bm{\theta}_{S^{(1)},S^{(2)}})$ is written as the complete
likelihood integrated over the latent variables.

Let $Z^m = (Z_{i}^m)_{i=1,\dots,n_1^m}$ denote the set a latent variables
encoding the clustering of the row nodes in network $m$. Similarly, we define
and $W^m = (W_{j}^m)_{j=1,\dots,n_2^m}$.
The log-likelihood of the observations $\bm{X}=(X^1, \dots, X^M)$ is written as
the complete likelihood integrated over the latent variables $(\bm{Z},\bm{W}) =
    (Z^m,W^m)_{m=1 ,\dots,M}$:
\begin{equation}
    \label{eq:log-likelihood}
    \ell(\bm{X};\bm{\theta}_{S^{(1)},S^{(2)}}) = \sum_{m=1}^M \log \sum_{ Z ^m ,W^m} \exp\{\ell(X^m|Z^m, W^m; \bm{\alpha}) + \ell(Z^m;\bm{\pi}) + \ell(W^m;\bm{\rho})\}
\end{equation}
where
\begin{align*}
    \ell(X^m|Z^m, W^m; \bm{\alpha}) = \sum_{i=1}^{n_1^m}\sum_{j=1}^{n_2^m} \sum_{q\in\mathcal{Q}_1^m}      & \sum_{r\in\mathcal{Q}_2^m} \mathbbb{1}_{Z_{i}^m=q, W_{j}^m=r} \log f(X_{ij}^m; \alpha_{qr}),                               \\
    \ell(Z^m;\bm{\pi}) = \sum_{i=1}^{n_1^m}\sum_{q\in\mathcal{Q}_1^m} \mathbbb{1}_{Z_{i}^m=q} \log \pi_q^m & ,~\text{and}~~ ~ \ell(W^m;\bm{\rho}) = \sum_{j=1}^{n_2^m} \sum_{r\in\mathcal{Q}_2^m}\mathbbb{1}_{W_{j}^m=r} \log \rho_r^m. \\
\end{align*}
where  $f$ is defined in~\Cref{eqn:lbm-emission}.
We obtain the log-likelihood of the iid-colBiSBM model by setting
$\bm{\pi}^m=\bm{\pi}$ and $\bm{\rho}^m=\bm{\rho}, \forall m$,  $\mathcal{Q}_1^m = \{1, \dots Q_1\}$  and $\mathcal{Q}_2^m = \{1, \dots Q_2\}$. For the $\pi$-colBiSBM and
$\rho$-colBiSBM models we obtain the log-likelihood by fixing either the row or
column proportions, and by setting the relevant support matrix to the matrix
full of ones.

\subsection{Identifiability of the models}
\label{ssec:identifiability-of-the-models}

We prove the identifiability of the parameters, namely that if
$\ell(\bm{X};\bm{\theta}) = \ell(\bm{X};\bm{\theta}')$ for any collection
$\bm{X}$ then $\bm{\theta} = \bm{\theta}'$. The identifiability of the
parameters of the simple biSBM is demonstrated in
\cite{keribinEstimationSelectionLatent2015} as an extension of the result
of~\cite{celisseConsistencyMaximumlikelihoodVariational2012}.
\cite{chabert-liddellLearningCommonStructures2024} gave identifiability
conditions for collections of simple (non bipartite) networks. We provide here
conditions which ensure the identifiability of the model. Note that the
conditions are sufficient but not necessary.

In $\pi\rho$-colBiSBM, we allow the block proportions to be null in certain
networks. Let us first define the restricted parameters for each network,
corresponding to the non-null block proportions:
\begin{eqnarray*}
    \overline{\bm{\pi}}^m &=& \left({\pi}_q^m\right)_{q \in \mathcal{Q}_1^m}, \quad \quad
    \overline{\bm{\rho}}^m = \left({\rho}_r^m\right)_{r \in \mathcal{Q}_2^m}\\
    \overline{\bm{\alpha}}^m &=& \left({\alpha}_{qr}\right)_{q \in \mathcal{Q}_1^m, r \in \mathcal{Q}_2^m}
\end{eqnarray*}

\begin{property}
    \label{prop:identifiability-iid-pirho}
    {\normalfont }
    The parameters $\bm{\theta} = (\bm{\pi}, \bm{\rho}, \bm{\alpha})$ of the iid-colBiSBM model
    are identifiable up to a
    label switching of the blocks under the following  conditions:
    \begin{enumerate}[label=(1.\arabic*)]
        \item $\exists m^*\in\{1,\dots,M\} : n^{m^*}_1 \geq 2 Q_2 - 1~\text{and}~n^{m^*}_2 \geq 2 Q_1 - 1$.\label{item:iid-n}
        \item The coordinates of vector $\bm{\alpha} \bm{\rho}$ are
              distinct.\label{item:iid-tau1}
        \item The coordinates of vector $\bm{\pi}^{\top}\bm{\alpha}$ (where $\bm{\pi}^{\top}$
              is the transpose of $\bm{\pi}$) are distinct.\label{item:iid-tau2}
    \end{enumerate}%

    {\normalfont} The parameters
    $\bm{\theta}_{S^{(1)}, S^{(2)}} = (\bm{\pi}^1, \dots, \bm{\pi}^M, \bm{\rho}^1, \dots, \bm{\rho}^M, \bm{\alpha})$ of the $\pi\rho$-colBiSBM
    are identifiable up to a label switching of the blocks if the following
    conditions are met:
    \begin{enumerate}[label=(2.\arabic*)]
        \item $\forall m\in\{1,\dots M\} : n^{m}_1 \geq 2 Q_2^{m} - 1$ and $n^{m}_2 \geq 2 Q_1^{m} - 1$\label{item:pirho-size}
        \item $\forall m\in\{1,\dots,M\},$ the coordinates of the vector $\overline{\bm{\alpha}}^m\overline{\bm{\rho}}^m$ are distinct.\label{item:pirho-distinct-row}
        \item $\forall m\in\{1,\dots,M\},$ the coordinates of the vector $(\overline{\bm{\pi}}^m)^{\top}\overline{\bm{\alpha}}^m$ are distinct.\label{item:pirho-distinct-col}
        \item All entries of $\bm{\alpha}$ are unique.\label{item:pirho-alpha}
    \end{enumerate}
\end{property}

The proof is provided
in\ifNOSUPP~supplementaries.\else
~\Cref{sec:proof-of-the-idenfiability-result}\fi.

\subsection{Variational estimation of the parameters}
\label{ssec:variational-estimation-of-the-parameters}

We aim to estimate $\bm{\theta}_{S^{(1)}, S^{(2)}}$ by maximizing the
log-likelihood of ~\Cref{eq:log-likelihood} for fixed numbers of blocks
$(Q_1,Q_2)$ and fixed support matrices $(S^{(1)}, S^{(2)})$ in the case of the
$\pi\rho$-colBiSBM model. However, in practice, this log-likelihood is not
tractable as it requires to sum over $\sum_{m=1}^{M} (Q_1^{m})^{n_1^m}
    (Q_2^{m})^{n_2^m}$ terms. A reliable method to tackle this problem is to rely
on a variational version of the Expectation Maximization (VEM) algorithm
(\cite{govaertBlockClusteringBernoulli2008,daudinMixtureModelRandom2008}).

The maximization of the log-likelihood is replaced by the maximization of a
variational lower bound of the log-likelihood of the observed data. For ease of
reading, the indices of $\bm{\theta}_{S^{(1)}, S^{(2)}}$ are dropped and we use
the short notation $\bm{\theta}$. Let $\mathcal{R}$ be an approximation of
$p(\bm{Z},\bm{W}|\bm{X} ;\bm{\theta})$. Then:
\begin{eqnarray}\label{eq:Lowerbound1}
    \ell(\bm{X};\bm{\theta}) &\geq& \ell(\bm{X};\bm{\theta}) -  {\displaystyle D_{\text{KL}}(\mathcal{R} \parallel p(\bm{Z},\bm{W}|\bm{X}
    ;\bm{\theta}))} \nonumber \\
    &=&  \mathbb{E}_{\mathcal{R}}[\ell(\bm{X},\bm{Z},\bm{W};\bm{\theta})] -
    \mathcal{H}(\mathcal{R}(\bm{Z},\bm{W}))\\
    &\eqcolon& 	\mathcal{J}(\mathcal{R};\bm{\theta}) \nonumber
\end{eqnarray}
where $ D_{\text{KL}}$ is the Kullback-Leibler divergence and $\mathcal{H} $ is the entropy. $\mathcal{J}(\mathcal{R};\bm{\theta}) $ is known as the evidence lower bound (ELBO). Note that  	$\mathcal{J}(\mathcal{R};\bm{\theta})=\ell(\bm{X};\bm{\theta})$ if $\mathcal{R} = p(\bm{Z},\bm{W}|\bm{X} ;\bm{\theta})$. So, the objective is now to maximize $\mathcal{J}(\mathcal{R};\bm{\theta}) $  with respect to $\bm{\theta}$ and $\mathcal{R}$. This task can be performed thanks to the reformulation of the lower bound provided in Equation \eqref{eq:Lowerbound1}.
The VEM algorithm is an iterative process aimed at maximizing the lower bound,  that alternates between the E-step
and the M-step. During the E-step $\mathcal{J}(\mathcal{R};\bm{\theta})$ is
maximized w.r.t. $\mathcal{R}$ and for a current value of $\bm{\theta}$, resulting into a minimization of  ${\displaystyle D_{\text{KL}}(\mathcal{R} \parallel p(\bm{Z},\bm{W}|\bm{X}	;\bm{\theta}))}$, which can not be performed directly.
The M-step maximizes
$\mathcal{J}(\mathcal{R};\bm{\theta})$ w.r.t. $\bm{\theta}$ for a given
variational distribution $\mathcal{R}$.
$\mathcal{R}$ must be chosen so that the quantity $\mathbb{E}_{\mathcal{R}}[\ell(\bm{X},\bm{Z},\bm{W};\bm{\theta})] $ can be calculated explicitly. By conditional independence of the row and column latent variables given the observations and by independence of the networks we have:
\begin{eqnarray*}
    p(\bm{Z},\bm{W}|\bm{X}
    ;\bm{\theta}) &=& p(\bm{Z}|\bm{X}
    ;\bm{\theta})p(\bm{W}|\bm{X}
    ;\bm{\theta})\\
    &=& \prod_{m=1}^M p(Z^m |X^m
    ;\bm{\theta})p(W^m|X^m
    ;\bm{\theta}).
\end{eqnarray*}
$\mathcal{R}(\bm{Z},\bm{W})$ is assumed to respect the same independence structure. Moreover, following~\cite{govaertBlockClusteringBernoulli2008,daudinMixtureModelRandom2008}, $\mathcal{R}(\bm{Z})$ and $\mathcal{R}( \bm{W})$ are chosen as product distributions, thus neglecting the conditional dependencies between nodes of a given network. Thus, $p(Z^m |X^m;\bm{\theta})$   is approximated by  $\mathcal{R}(Z^m) = \prod_{i=1}^{n_m}p_{\mathcal{R}}(Z^m_{i})$   which is parametrized by the  probabilities $\tau^{1,m}_{iq}$.
$$	\tau^{1,m}_{iq} = \mathbb{P}_{\mathcal{R}}(Z^m_{i} = q).  $$
The same holds for $p(W^m |X^m;\bm{\theta})$ and we set:
$$
    \tau^{2,m}_{jr} = \mathbb{P}_{\mathcal{R}}(W^m_{j} = r).
$$
These quantities are the approximations of the posterior node clustering  probabilities. In the following we identify $\mathcal{R}$ to these quantities.

For the colBiSBM models, the VE-step is computed independently for each network
and during the M-step the parameters are updated with formulas that link the
networks together.

We describe below the update procedure to go from step $t$ to $t+1$, in our
implementation this is done until convergence and we use a mini-batching
approach as it converges quicker in the experiments. To initialize the
algorithm one can do a spectral clustering of the nodes to determine the first
posterior node clustering probabilities.

\paragraph{Variational E step}
\label{ssec:variational-e-step}

At this step we maximize with respect to the variational distribution
$\mathcal{R}$: \[
    \widehat{\mathcal{R}}^{(t+1)} = \argmax_{\mathcal{R}}
    \mathcal{J}(\mathcal{R},\bm{\widehat{\theta}}^{(t)}).
\]

And we obtain the following formulas for the $\bm{\tau}^m =
    ((\tau_{iq}^{1,m})_{\substack{i=1,\dots,n_1^m\\q \in \mathcal{Q}_1^m}},
    (\tau_{jr}^{2,m})_{\substack{j=1,\dots,n_2^m\\r\in \mathcal{Q}_2^m}})$:


\begin{equation*}
    \begin{cases}
        \widehat{\tau}_{iq}^{1,m} \propto \widehat{\pi}_{q}^{m,(t)} \prod_{j=1}^{n_2^m}\prod_{r\in\mathcal{Q}_2^m} f(X_{ij}^m;\widehat{\alpha}_{qr}^{(t)})^{\widehat{\tau}_{jr}^{2,m,(t+1)}}  & \forall i = 1, \dots , n_1^m, q \in \mathcal{Q}_1^m \\
        \widehat{\tau}_{jr}^{2,m} \propto \widehat{\rho}_{r}^{m,(t)} \prod_{i=1}^{n_1^m}\prod_{q\in\mathcal{Q}_1^m} f(X_{ij}^m;\widehat{\alpha}_{qr}^{(t)})^{\widehat{\tau}_{iq}^{1,m,(t+1)}} & \forall j = 1, \dots , n_2^m, r \in \mathcal{Q}_2^m
    \end{cases}
\end{equation*}

which are used to update iteratively the values.

\paragraph{M step}
\label{ssec:m-step-of-the-algorithm}
At iteration $(t)$ the M-step maximizes the variational bound with respect to
the model parameters $\bm{\theta}$:
\[
    \widehat{\bm{\theta}}^{(t+1)} = \arg \max_{\bm{\theta}} \mathcal{J}(\mathcal{\widehat{\mathcal{R}}}^{(t+1)},\bm{\theta})
\]
The following quantities are involved in the obtained formulas:
\begin{align*}
    e^{m}_{qr} = \sum_{i=1}^{n_1^m} \sum_{j=1}^{n_2^m} \tau_{iq}^{1,m} \tau_{jr}^{2,m} X_{ij}^m
     & , & n^{m}_{qr} = \sum_{i=1}^{n_1^m} \sum_{j=1}^{n_2^m} \tau_{iq}^{1,m} \tau_{jr}^{2,m}
     & , & n^{1,m}_{q} = \sum_{i=1}^{n_1^m} \tau_{iq}^{1,m}
     & , & n^{2,m}_{r} = \sum_{j=1}^{n_2^m} \tau_{jr}^{2,m}
\end{align*}
The block proportions estimations depend on the model we consider, $$
    \begin{array}{ccccccccl}
        \widehat{\pi}_q     & = & \frac{\sum_{m=1}^{M} n^{1,m}_{q}}{\sum_{m=1}^{M} n_1^m} & , & \widehat{\rho}_r     & = & \frac{\sum_{m=1}^{M} n^{2,m}_{r}}{\sum_{m=1}^{M} n_2^m} & \text{for } \text{iid-colBiSBM}      \\
        \widehat{\pi}_q^{m} & = & \frac{n^{1,m}_{q}}{n_1^m}                               & , & \widehat{\rho}_r^{m} & = & \frac{n^{2,m}_{r}}{n_2^m}                               & \text{for } \pi\rho\text{-colBiSBM}.
    \end{array}
$$
while the connectivity parameters $\alpha_{qr}$ are estimated as the ratio of
the number of observed interactions between row block $q$ and column block $r$
among all networks over the number of possible interactions:
\begin{align*}
    \widehat{\alpha}_{qr} = \frac{\sum_{m=1}^{M} e^{m}_{qr}}{\sum_{m=1}^{M} n^{m}_{qr}}.
\end{align*}

\paragraph{Output of the algorithm}
The inference algorithm supplies for each node of any network its probability
of clustering given its connections namely the $\widehat{\tau}_{iq}^{1,m} $ and
$\widehat{\tau}_{jr}^{2,m}$. These quantities can be used to perform hard
clustering (taking the \emph{maximum a posteriori}, in short, MAP) and so to
reorganize the bi-adjacency matrices. Note that contrarily to sep-biSBM
(\Cref{sec:data-motivation-and-bipartite-stochastic-block-model}), the
clustering we obtain here are coherent across the networks. In other words, two
species from networks $m$ and $m'$ affected to a given cluster $q$ do play the
same role in their respective ecosystem. On the contrary, from the ecological
point of view, one can check if a given species represented by two different
nodes in networks $m$ and $m'$ is affected to the same block or plays a
different role in the two ecosystems.


\subsection{Selecting the numbers of blocks $(Q_1,Q_2)$}
\label{ssec:model-selection}\label{sssec:q-1-and-q-2-block-number-selection}
\Cref{ssec:variational-estimation-of-the-parameters} was dedicated to the estimation of  $\bm{\theta}$ for fixed  $(Q_1,Q_2)$. In practice the numbers of blocks need to be inferred. We propose a penalized likelihood criterion.
We first consider the iid-colBiSBM  and then the  $\pi\rho$-colBiSBM which requires a few more work due to the possibly null block proportions in some of the networks.

\subsubsection{A penalized criterion adapted to collections of networks}
\label{ssec:a-penalized-criterion-adapted-to-collections-of-networks}
The Integrated Classification Likelihood (ICL), introduced by~\cite{biernackiAssessingMixtureModel2000,
    daudinMixtureModelRandom2008},  is a well-established tool for selecting the appropriate number of blocks of SBM and biSBM.  The ICL is derived from an asymptotic
approximation of the marginal complete likelihood  where the parameters are integrated out using a prior distribution, resulting in a
penalized likelihood criterion. Usage has shown good recovery properties for
the number of blocks. The generic expression of the ICL is
\[
    \text{ICL} = \max_{\bm{\theta}} \mathbb{E}_{\widehat{\mathcal{R}}} [\ell(\bm{X,Z,W;\theta})] - \frac{1}{2}\text{pen}
\]
where $\text{pen}$ penalizes the complexity of the model. ICL is known to
encourage well-separated blocks by imposing a penalty on the entropy of node
grouping.
However, the objective of our approach extends beyond grouping nodes into
coherent blocks. We also aim to assess the similarity of connectivity patterns
across different networks. Consequently, we suggest to permit models that offer
more flexible node grouping by not penalizing on entropy. This leads us to
formulate a BIC-like criterion in the following manner:
\[
    \text{BIC-L} = \max_{\bm{\theta}} \mathbb{E}_{\widehat{\mathcal{R}}} [\ell(\bm{X,Z,W;\theta})] + \mathcal{H(\widehat{R})} - \frac{1}{2}\text{pen} = \max_{\bm{\theta}} \mathcal{J(\widehat{R}, \bm{\theta})} - \frac{1}{2}\text{pen}
\]
We provide below the penalty expressions for iid-colBiSBM and
$\pi\rho$-colBiSBM. 

\subsubsection{Model selection for iid-colBiSBM}
\label{ssec:model-selection-for-iid-colbisbm}
For iid-colBiSBM, the penalty follows from the same logic as for the
classical biSBM since the networks are independent realizations of the same biSBM:
\[
    \text{BIC-L}(\bm{X},Q_1, Q_2) = \max_{\bm{\theta}}
    \mathcal{J} (\mathcal{\hat{R}}, \bm{\theta})
    - \frac{1}{2} [\text{pen}_{\pi}(Q_1) + \text{pen}_{\rho}(Q_2) +
        \text{pen}_{\alpha}(Q_1, Q_2)]
\]
with
\begin{align*}
    \text{pen}_{\pi}(Q_1) = (Q_1 - 1)\log\bigg(\sum_{m=1}^{M}n_{1}^{m}\bigg) & , &
    \text{pen}_{\rho}(Q_2) = (Q_2 - 1)\log\bigg(\sum_{m=1}^{M}n_{2}^{m}\bigg),
\end{align*}
\[
    \text{pen}_{\alpha}(Q_1, Q_2) = Q_1  Q_2 \log(N_M
    ),
\]
where
\begin{equation}
    \label{eq:max-nodes-interactions}
    N_M = \sum_{m = 1}^{M} n_{1}^{m}  n_{2}^{m},
\end{equation}
is the  number of entries in all the bi-adjacency matrices.
The terms $\text{pen}_{\pi}(Q_1)$ and $\text{pen}_{\rho}(Q_2)$ correspond to the estimation of the block
proportions. The term  $\text{pen}_{\alpha}(Q_1, Q_2)$ is the penalty for the connectivity.
And finally $Q_1$ and $Q_2$ are chosen as:
\[
    (\widehat{Q_1}, \widehat{Q_2}) = \argmax_{Q_1, Q_2} \text{BIC-L}(\bm{X}, Q_1, Q_2).
\]

\subsubsection{Model selection for $\bf{\pi\rho}$-colBiSBM}
\label{ssec:model-selection-for-pi-rho-colbisbm}
In model $\bf{\pi\rho}$-colBiSBM, we need to choose not only $(Q_1, Q_2)$ but also the support matrices $(S^{(1)}, S^{(2)})$. The penalty term must take into account the dimension of the supports $(S^{(1)}, S^{(2)})$ which is the number of non-null proportions namely $Q_{1}^m$ and $Q_{2}^m$ defined in Equation \eqref{eq:dimS1S2}.
Considering the same principle as ICL described before, the penalty derives
from a prior distribution on $(S^{(1)}, S^{(2)})$: a uniform prior over
$\{1,\dots, Q_1\} \times \{1,\dots, Q_2\}$ is set as a prior on $(Q_1^{m},
    Q_2^{m})$; then conditionally to $(Q_1^{m}, Q_2^{m})$, the place of the
non-empty blocks in network $m$ (i.e., $(S_{mq}^{(1)})_{q=1,\dots Q_1}$ and
$(S_{mr}^{(2)})_{r=1,\dots Q_2}$) are uniformly distributed, resulting into a
penalty of the form
\[
    \log p_{Q_1}(S^{(1)}) = - M \log(Q_1) - \sum_{m=1}^{M} \log {Q_1 \choose Q_1^{m}}
\]
where ${Q_1 \choose Q_1^{m}}$ is the number of ways to place the $Q_1^{m}$ non
empty blocks among the $Q_1$ possible blocks. The same formula holds for
$p_{Q_2}(S^{(2)})$. Using the Laplace approximation and the prior distributions
we obtain the following criterion:
\[
    \text{BIC-L}(\bm{X},Q_1, Q_2) =
    \max_{S^{(1)},S^{(2)}} \bigg[
    \max_{\bm{\theta}_{S^{(1)},S^{(2)}}} \mathcal{J}(\mathcal{\hat{R}},
    \bm{\theta}_{S^{(1)},S^{(2)}})
    - \frac{\text{pen}(Q_1,Q_2,S^{(1)},S^{(2)},\alpha,\pi,\rho)}{2}\bigg].
\]
The penalty is composed of the following terms:
\[
    \text{pen}_{\pi}(Q_1, S^{(1)})  + \text{pen}_{\rho}(Q_2, S^{(2)}) + \text{pen}_{\alpha}(Q_1, Q_2, S^{(1)}, S^{(2)}) + \text{pen}_{S^{(1)}}(Q_1) + \text{pen}_{S^{(2)}}(Q_2),
\]
with
\begin{align*}
    \text{pen}_{S^{(1)}}(Q_1) = -2 \log p_{Q_1} (S^{(1)}) & , &
    \text{pen}_{S^{(2)}}(Q_2) = -2 \log p_{Q_2} (S^{(2)}),
\end{align*}
\[ \text{pen}_{\pi}(Q_1, S^{(1)}) = \sum_{m=1}^{M} (Q_{1}^{m} - 1)
    \log n_{1}^{m},
    ~\text{pen}_{\rho}(Q_2, S^{(2)}) = \sum_{m=1}^{M} (Q_{2}^{m} - 1)
    \log n_{2}^{m},
\]
and
\[
    \text{pen}_{\alpha}(Q_1, Q_2, S^{(1)}, S^{(2)}) = \left(\sum_{q=1}^{Q_1}
    \sum_{r=1}^{Q_2} \mathbbb{1}_{((\supp{1})'\supp{2})_{qr} > 0}\right) \log (N_M),
\]
where $N_M$ is defined in~\Cref{eq:max-nodes-interactions}. $(Q_1, Q_2)$ are
finally chosen by maximizing the BIC-L criterion:
\[
    (\widehat{Q_1}, \widehat{Q_2}) = \argmax_{Q_1, Q_2} \text{BIC-L}(\bm{X}, Q_1, Q_2).
\]

\subsubsection{Practical model selection}
\label{sssec:practical-model-selection}
The choice of $(Q_1,Q_2)$ and the estimation of the model  parameters are computationally expensive. All the models with possible values of $(Q_1,Q_2)$ should be fitted and
compared through the $\text{BIC-L}(\bm{X}, Q_1, Q_2)$ criterion. In addition, for a fixed value of $(Q_1,Q_2)$,  the variational EM being sensitive to the initialization,
multiple runs are needed. 


We propose a partial exploration of the model space by a split and merge
approach. The first step is a greedy exploration to find an estimation of the
mode of the criterion. The second step is a moving window centered around the
first mode to refine the first estimation by leveraging all the previously
fitted models. The procedure, detailed in~\Cref{alg:model-selection} is
implemented in the \texttt{R}-package \textsf{colSBM}.


\SetKwComment{Comment}{$\triangleleft$~}{}
\begin{algorithm}
    \SetKwInOut{Input}{Tuning parameters}
    \caption{Model selection algorithm}\label{alg:model-selection}
    \KwData{$\bm{X}$ a collection of bipartite networks}
    \Input{$max_{\text{ge}}$,
        $max_{\text{mw}}$ and $depth$}

    \Begin{
        - Fit colBiSBM model with $\bm{Q} = (1,2)$ and $\bm{Q} = (2,1)$
    }

    $step_{\text{ge}} \gets 0$\\
    \For{$\bm{Q}_{\text{curr}}\in \{(1,2),(2,1)\}$}{
    \While{$\BICL$ increases and $step_{\text{ge}} \leq max_{\text{ge}}$ \Comment*[r]{Greedy exploration}}{

    - Fit the models for $\bm{Q} \in \bm{Q}_{\text{curr}} + \{(1,0), (0,1)\}$ \Comment*[r]{Splits}
    \If{$Q_{\text{curr}}$ allows to merge}{
    - Fit the models for $\bm{Q} \in \bm{Q}_{\text{curr}} - \{(1,0), (0,1)\}$ \Comment*[r]{Merges}
    }
    - $\bm{Q}_{\text{curr}} \gets \argmax \BICL(\bm{X}, Q_1, Q_2)$ and
    $step_{\text{ge}} \gets step_{\text{ge}} + 1$\\
    }
    }
    - $\bm{Q}_{\text{ge}} \gets \bm{Q}_{\text{curr}}$ and
    $step_{\text{mw}} \gets 0$\\
    \While{$\BICL$ increases and $step_{\text{mw}} \leq max_{\text{mw}}$ \Comment*[r]{Moving window}}{
    \For{$\bm{Q}_{\text{curr}} = \bm{Q}_{\text{curr}} - depth, ...,
        \bm{Q}_{\text{curr}} + depth$\Comment*[r]{Forward pass}}{
    - Fit the models by splitting from $\bm{Q}_{\text{curr}}$\\
    }
    \For{$\bm{Q}_{\text{curr}} = \bm{Q}_{\text{curr}} + depth, ...,
        \bm{Q}_{\text{curr}} - depth$\Comment*[r]{Backward pass}}{
    - Fit the models by merging from $\bm{Q}_{\text{curr}}$\\
    }
    - $\bm{Q}_{\text{curr}} \gets \argmax \BICL(\bm{X}, Q_1, Q_2)$
    and $step_{\text{mw}} \gets step_{\text{mw}} + 1$\\
    }
    \KwRet{$\widehat{Q_1}, \widehat{Q_2} = \argmax \BICL(\bm{X}, Q_1, Q_2)$ with
        $\widehat{\bm{\theta}}, \widehat{\bm{Z}}, \widehat{\bm{W}}$ and
        $\widehat{S}^{(1)}, \widehat{S}^{(2)}$ for $\pi\rho$-colBiSBM models.}
\end{algorithm}


\subsection{Searching for   common connectivity structures among a collection of networks}
\label{sssec:testing-common-connectivity-structure}

\paragraph{ sep-biSBM versus colBiSBM}  Beyond determining the optimal number of blocks, the BIC-L criterion we derived
can also be used to test whether two or more bipartite networks share a common
mesoscopic structure. To do so, the first model comparison that we may perform
is between the colBiSBM models and the sep-biSBM model to determine if the
networks share a common structure. We decide that networks share a common
structure if:
\[
    \max_{Q_1, Q_2} \text{BIC-L}_{colBiSBM}(\bm{X}, Q_1, Q_2) > \sum_{m=1}^{M}\max_{Q_1^{m}, Q_2^{m}} \text{BIC-L}_{biSBM}(\bm{X}^m, Q_1^{m}, Q_2^{m}).
\]

\paragraph{Partition of a collection of networks}
At a more advanced stage, when considering a large collection of networks as
the one described in~\cite{doreRelativeEffectsAnthropogenic2021}, one may seek
to identify subgroups of networks exhibiting similar connectivity patterns. In
other words, the goal is to perform a network partitioning procedure based on
their mesoscopic structure i.e., find $ \mathcal{G} = (\mathcal{M}_g)_{g=1,
            \dots, G}$ a partition of $\{1,\dots, M\}$ such that
\[
    ~\forall m \in \mathcal{M}_g,
    ~X^m \sim
    \mathcal{F}\text{-biSBM}(Q_1^{g}, Q_2^{g}, \bm{\pi}^m, \bm{\rho}^m, \bm{\alpha}^g)
    , \]
where $\bm{\alpha}^g$ depends on the group the network belongs to. For any
partition $\mathcal{G}$ , we define an adequacy score, denoted by
$Sc(\mathcal{G})$, which quantifies the fit of the partition according to the
BIC-L criterion:
\begin{equation}\label{eq:BICLPartitioning}
    Sc(\mathcal{G}) = \sum_{g=1}^{G} \max_{Q_1^{g}, Q_2^{g}} \text{BIC-L}((X^m)_{m\in\mathcal{M}_g},Q_1^{g}, Q_2^{g})
\end{equation}
Thus the score consists of the sum of the BIC-L computed on the sub-collections
for the partition $\mathcal{G}$. $B_M$ (Bell number) possible partitions have to be compared
though this score. If $M$ is small, the scores of the $B_M$ possible partitions
can be computed by fitting a colSBM to any of the $2^M$ subcollections of networks. However,
if $M$ is large, this strategy is not realistic anymore for computational
considerations. In this case, one can resort to a hierarchical research strategy. In a few
words, adjust a unique model on the full collection and then proceed by sequential
divisions of the partition based on dissimilarities between parameters.  Denoting
$\mathcal{G}^*$ the new partition obtained from $\mathcal{G}$, if $Sc(\mathcal{G}^{*}) > Sc(\mathcal{G})$
try to partition further, else stop and return $\mathcal{G}$.
For extensive details see\ifNOSUPP~supplementaries\else~\Cref{sec:details-of-the-network-partitioning-algorithm}\fi.

\subsection{Accuracy of the inference strategy on synthetic datasets}
\label{ssec:simulation-study}
We illustrate the accuracy of our inference method on synthetic datasets.  The results of our extensive simulation study is available in supplementary material.
In \ifNOSUPP~the first Section of our numerical experiments\else~\Cref{ssec:efficiency-of-the-inference}\fi, we prove the capacity of our inference procedure to recover the number of blocks. The  corresponding nodes  clustering is assessed under the different colBiSBMs.
The ability to choose the correct colBiSBM (through BIC-L) is tested in\ifNOSUPP~the second Section\else~\Cref{ssec:from-fixed-block-proportions-to-varying-block-proportions}\fi.
The interest of a joint learning under a colBiSBM compared to fitting separated biSBMs is demonstrated in\ifNOSUPP~a third Section\else~\Cref{ssec:information-transfer-between-networks}\fi.
Finally, the efficiency of the partitioning procedure of a collection of networks is investigated in\ifNOSUPP~the fourth Section\else~\Cref{ssec:clustering-of-simulated-networks}\fi.
All these simulations are reproducible, the code is available at \url{https://github.com/Polarolouis/code-colbisbm/}.

\section{Analyzing  real-world ecological networks}
\label{sec:applications-on-ecological-networks}

To demonstrate the practical relevance of our approach, we apply it to
real-world networks obtained from~\cite{baldockSystemsApproachReveals2019}
(presented in~\Cref{ssec:data-motivation-and-bipartite-stochastic-block-model})
and~\cite{baldockDailyTemporalStructure2011}.



In a preliminary study (results not shown here), we first fitted the various
colBiSBM models on the collection constituted of the four British networks
introduced in~\Cref{ssec:data-motivation-and-bipartite-stochastic-block-model}.
We obtained the following results: on this collection the iid-colBiSBM was
selected as the best model, confirming the fact that they share a common
organization. In that particular case, the assumption of equal block
proportions is adapted to the data. To strengthen our evaluation, we augmented
the collection with a savanna pollination network deduced from the data
collected in Kenya and presented in~\cite{baldockDailyTemporalStructure2011}.
Although the original dataset focused on the daily temporal structure, we
concatenated the data in a single merged network.


\subsection{Joint estimation and  partitioning  of the five Baldock networks}
\label{ssec:joint-estimation-and-clustering-of-baldock-networks}


We applied our partitioning method on those five networks, using the four
models. The BIC-L obtained for the best partition and the various joint
modeling are presented in~\Cref{tab:baldock-bicl}.

\begin{table}[H]
    \caption{\label{tab:baldock-bicl}BICL values for different models on the~\cite{baldockSystemsApproachReveals2019,baldockDailyTemporalStructure2011} datasets.}
    \centering
    \begin{tabular}[t]{|l|r|}
        \hline
        Model                              & BICL ($\uparrow$)    \\
        \hline
        iid-colBiSBM (Kenya and 4 British) & $\mathbf{-9466.866}$ \\
        \hline
        iid-colBiSBM on all networks       & $-9479.636$          \\
        \hline
        $\pi\rho$-colBiSBM on all networks & $-9497.920$          \\
        \hline
        Separate biSBM                     & $-9620.375$          \\
        \hline
    \end{tabular}
\end{table}

\Cref{tab:baldock-bicl} indicates
that all our models beat the separate biSBM (last line) hence there is an advantage to
model jointly those networks. The fact that the iid-colBiSBM on all networks is preferred
to the $\pi\rho$-colBiSBM can be explained by the cost of all the additional
free parameters for the $\pi\rho$ that hinder its BIC-L. Finally, the preferred
configuration is to split the networks in two partitions, one for the Kenyan
network and another for the British networks.
We can now interpret the results of the partitions obtained by the iid-colBiSBM and $\pi\rho$-colSBM models.

\begin{figure}[H]
    \centering
    \includegraphics[width=.8\textwidth,origin=c]{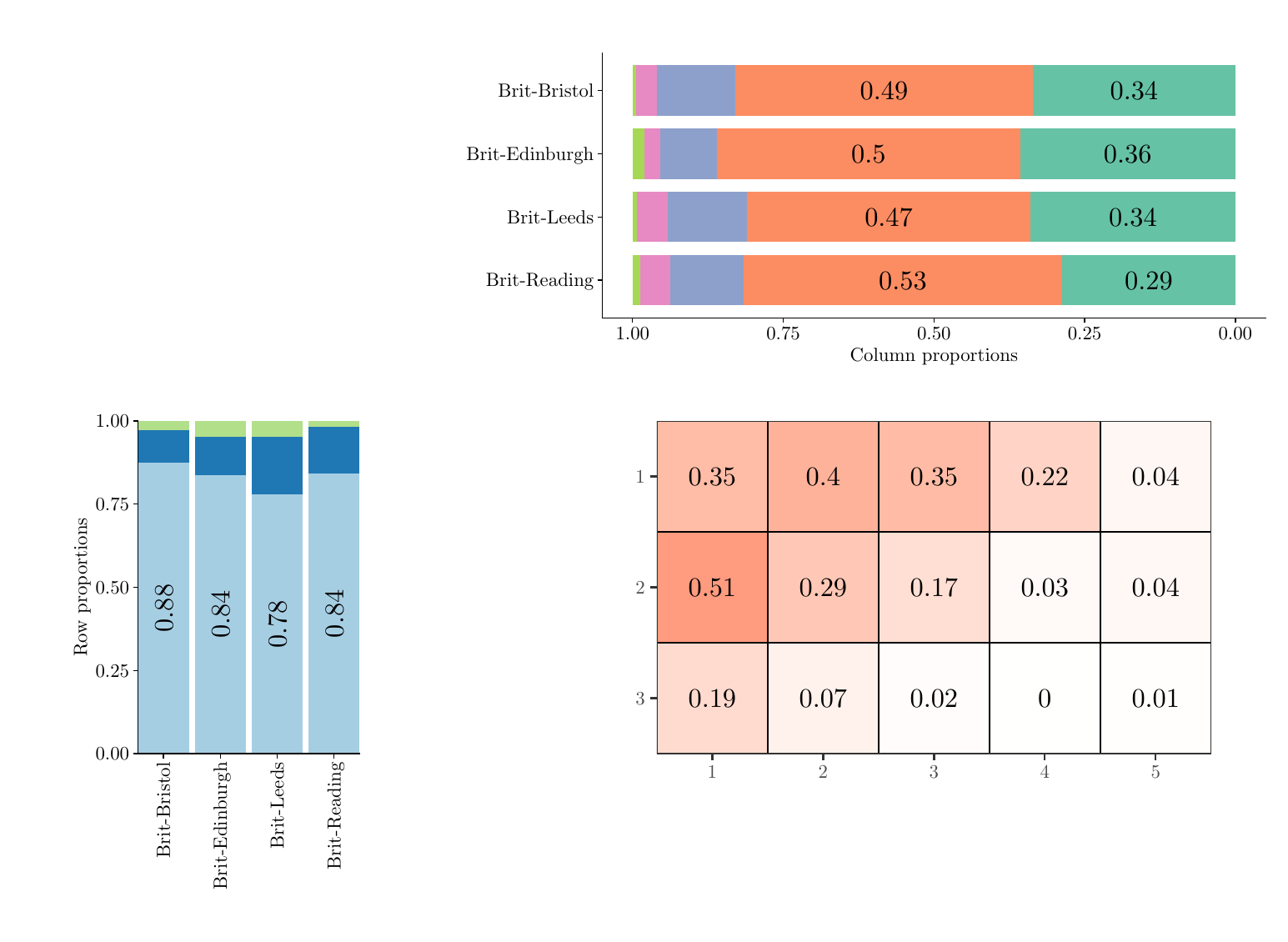}
    \caption{Result of an iid-colBiSBM clustering applied to the five Baldock networks, four British and one Kenyan.
        Left is Kenyan network structure, right is shared British structure.\label{fig:baldock-iid-clustering}}
\end{figure}

First, under the iid-colBiSBM, the clustering criterion (see
\Cref{sssec:testing-common-connectivity-structure}) prefers to separate the
Kenyan network from the British ones. Once partitioned, the method recovers a
structure commonly found in ecological networks: nestedness in both
collections. In the Kenyan network, it identifies two groups of pollinators
(row nodes) and two groups of plants (column nodes). The resulting connectivity
matrix displays a reverse-stair pattern, with generalist species in both rows
and columns interacting strongly among themselves (darker red shades,
$\alpha_{1,1} = 0.5$), while specialist species interact primarily with
generalists ($\alpha_{2,1}$ and $\alpha_{1,2}$) and show minimal interaction
with one another ($\alpha_{2,2} = 0.02$). In terms of block sizes, the
generalist species—although responsible for most of the interactions—form very
small fractions of the network: only about 2\% of the pollinators and 5\% of
the plants.

The structure for the British Baldock networks, shown
in~\Cref{fig:baldock-iid-clustering}, also exhibits nestedness with three
groups of pollinators and five groups of plants. Interestingly the generalists
connectivity coefficient is not the highest one, the highest one being with the
second pollinator group. The generalist pollinators are distributed across
blocks 1 and 2. Generalist plants are spread in 3 groups, 1, 2 and 3. When
comparing to~\Cref{fig:lbm-plant-pollinator}, the joint modeling allows to
refine the plant groups. In particular, it splits both generalists and
specialists adding two new functional groups. Further analysis with ecologists
may reveal the functional meaning of these new groups.

\begin{figure}[H]
    \centering
    \includegraphics[width=.8\textwidth,origin=c]{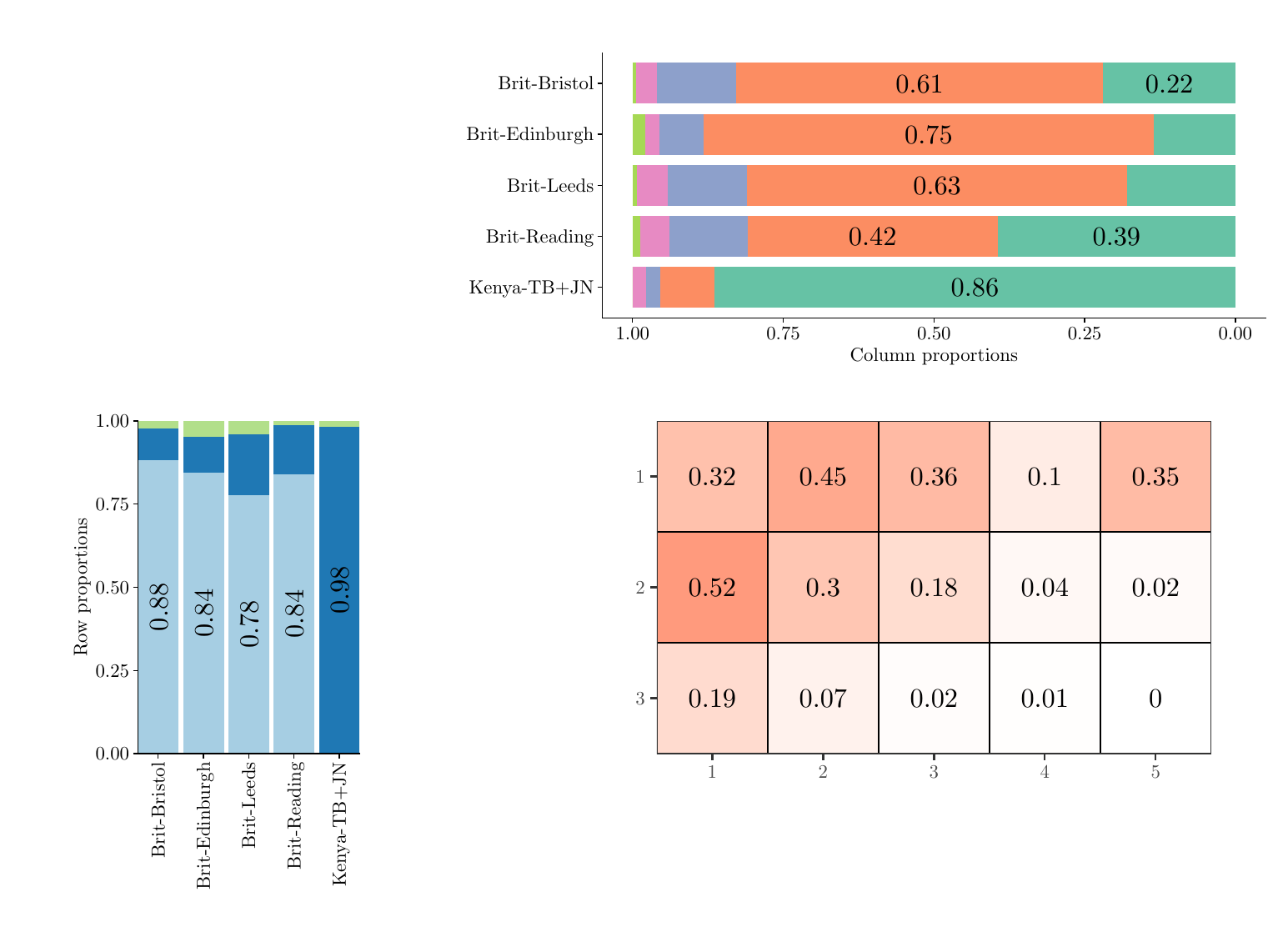}
    \caption{Result of a $\pi\rho$-colBiSBM clustering applied to the five Baldock networks, four British and one Kenyan.
        The shared structure for the five networks is in shades of red. The networks are in this order: Bristol, Edinburgh, Leeds, Reading and Kenyan\label{fig:baldock-pirho-clustering}}
\end{figure}

Secondly, we perform the network collection partition using the
$\pi\rho$-colBiSBM and plot the results in \Cref{fig:baldock-pirho-clustering}.
The flexibility induced by the varying block proportions leads to keeping
together the five networks (whereas they were separated using the iid-colBiSBM)
thus finding a shared structure between the Kenyan and British networks. Using
this approach, the pollinator blocks are left unchanged but the plant blocks
see at least a swap between blocks 4 and 5 along with some species of
\emph{former} block 4 joining \emph{former} block 5. The Kenyan network does
not populate the pollinator block 3 and sees most of its pollinators in block
2. For the plant blocks, it does not populate the first generalist block and
the trend is reversed when comparing to British networks: the British networks
populate block 4 and less block 5 whereas Kenyan network mainly ($86\%$) make
use of block 5 and not ($<10\%$) block 4. Note that, to go further, interesting
results could arise from comparing the nodes transfer between the two
structures seen
in~\Cref{fig:baldock-iid-clustering,fig:baldock-pirho-clustering}, since it
could inform us on functional role change of the blocks between the
clusterings. This ecological analysis is out of the scope of the present work.
\subsection{Edge prediction on ecological networks}
\label{sssec:transfer-on-ecological-data}

We now present a numerical experiment that demonstrates the benefits of joint
modeling for predicting edges. We compare the results of our model-based
approach to the performances obtained by a \emph{Variational Graph AutoEncoder}
(VGAE,~\cite{kipfVariationalGraphAutoEncoders2016a}). The objective is to prove
that joint modeling leverages information transfer.

To do so, we consider the four British networks
from~\cite{baldockSystemsApproachReveals2019} and randomly mask a proportion of
edges on one of the four networks, then we gauge the predicted interactions. We
use two kind of missing edges: missing links -- i.e, we replace a random number
of entries in the bi-adjacency matrix by a zero thus degrading the information
by potentially masking ones-- and missing dyads where we replace the value by a
\texttt{NA} allowing the model to take into account that the \enquote{value of} the
interactions are not known.

\def\hiddenlayerdim{4}
\def\latentdim{4}

We fit the four colBiSBM models on the degraded networks. We also fit a biSBM on the
degraded network: this plays the role of reference method where the information
of the other networks cannot be leveraged. In order to challenge our
probabilistic model based approach, we additionally fit a VGAE on the four
British networks jointly or on the altered network. In a few words, we use a
VGAE that takes as input features a non informative ordering of nodes as
integers. The encoder has three GraphSAGE
layers~(\cite{hamiltonInductiveRepresentationLearning2018}) of dimension
\hiddenlayerdim~ and the latent space is of dimension \latentdim. The output of
the encoder is two vectors $V^m_{1}$ (for pollinators) and $V^m_{2}$ (for
plants) for each network. $V^m_{1}$ (respectively $V^m_{2}$) dimensions are
$(n_1^m, \latentdim)$ (respectively $(n_2^m, \latentdim)$). For the decoder we
use an inner product decoder as proposed
by~\cite{kipfVariationalGraphAutoEncoders2016a} adapted for bipartite networks.
The predictions of the missing dyads or edges are done according to the fitting
models. For missing dyads, we predict the unobserved (\texttt{NA}) dyads. For
missing edges, we predict the altered edges. \\

\begin{figure}
    \centering
    \includegraphics[width=1\textwidth]{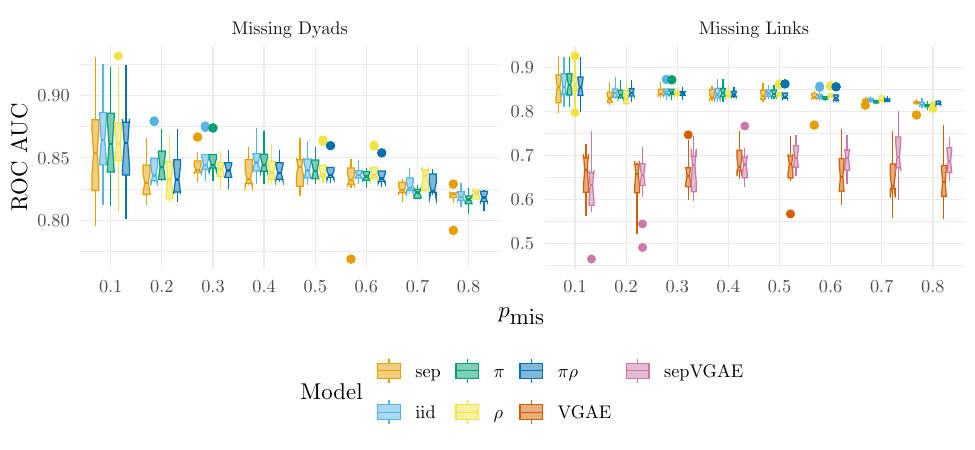}
    \caption{\textit{Information transfer on ecological data}: AUC of the recovered links in the missing dyads (left) and missing links (right). All the models are fitted on the Baldock British dataset with one of the networks altered. The \enquote{sep} model is a biSBM destined to serve as a reference.\label{fig:baldock-information-transfer}}
\end{figure}

In \Cref{fig:baldock-information-transfer}~ we plot the boxplots of the ROC AUC
metrics.
Our joint models perform marginally better than the sep-biSBM for the recovery of the
interactions thus indicating that the joint approach allow for information transfer
between the networks. The best models are iid-colBiSBM and $\pi$-colBiSBM returning the
higher AUC overall for the low $\pmis$.



From the two kinds of missing information, the missing dyad is the one that our
models tackle best. We explain this by the fact that missing interactions are
clearly labelled as \texttt{NA} and not taken into account to estimate the
parameters. The block parameter estimation are then only based on observed
interactions and nodes for which some interactions are missing in one network
can be estimated by using the information for the block in the other networks.
Whereas for the missing links experiment, the missing interactions are set as
$0$, potentially masking some interactions and being used to infer parameters,
degrading the performance of the model for interaction prediction.


Regarding the VGAE, we observe that VGAE trained jointly on the 4 networks does
not manage to transfer information. 
In both cases (sepVGAE and VGAE), the performances are far worse than the ones
of the probabilistic model approach.

The difference between the sep-biSBM and colBiSBM models are narrow but we
observe in the extensive simulation study that a nested structure (as the one
observed in these ecological networks) is usually well retrieved by a sep-biSBM
despite the alterations. However, for other network structures, we show that
the differences can be clearer (see
\ifNOSUPP~supplementaries\else~\Cref{ssec:information-transfer-between-networks}\fi).

\section{Discussion}
\label{sec:Discussion}


We introduce probabilistic models, namely colBiSBMs, that assume a shared
connectivity structure among a collection of networks. The models range from
the most constrained iid-colBiSBM to the more flexible $\pi\rho$-colBiSBM which
accommodates empty blocks and varying block proportions.
This joint modeling approach offers two main advantages compared to fitting
separate biSBMs. First, by leveraging a larger combined dataset, the joint
inference increases statistical power and can reveal finer structural patterns.
Second, it establishes explicit links between the blocks (node clusters) identified in
each network, allowing the corresponding clusters across networks to be
interpreted as functional groups that play analogous roles within their
respective ecosystems. This was illustrated in a dataset of five
plant-pollinator networks.


Note that a limitation of the iid-colSBM is that all the networks of the collection must have the same global density. 
The other models are a bit more flexible in this regard.
At present,
unlike~\cite{chabert-liddellLearningCommonStructures2024}, our models do not
incorporate a way to deal with common  structures up to a density variation.
Since there are many options to account for varying densities, we consider as a
promising direction for future work to generalize the model by allowing any
increasing function to mediate the transformation between network connectivity
structures. This would enable a colBiSBM variant capable of identifying
structural differences while explicitly accounting for density effects.
Concretely, such a formulation would associate with each network an increasing
function $d_m$ that transforms the shared connectivity matrix $\alpha$,
yielding the relation $\alpha^m = d_m(\alpha)$.


For the partition of a collection of networks,  the example we develop
in~\Cref{ssec:joint-estimation-and-clustering-of-baldock-networks} can be attained by an
exhaustive search (for 5 different elements there are 52 partitions). For larger collections, Bell
numbers grow too rapidly to consider exhaustively searching for the best
partition. We have then to resort to a partitioning algorithm like the one described in~\Cref{sssec:testing-common-connectivity-structure}. In our
simulation study (see\ifNOSUPP~in supplementaries\else~\Cref{ssec:clustering-of-simulated-networks}\fi), we assess its capacity to retrieve the true partition of the
collection of networks. However, the method reached its limits on the large
collection of~\cite{doreRelativeEffectsAnthropogenic2021}, presenting unstable
behaviors. This may be due to the fact that there are numerous connectivity
structures within this dataset and some of them are not as contrasted as in the
simulation study. Some perspectives proposed
by~\cite{rebafkaModelbasedClusteringMultiple2023a} could be adapted to bipartite
networks.

    \section*{Acknowledgements}
    \label{sec:acknowledgements}

    The authors are grateful to the INRAE MIGALE bioinformatics facility (MIGALE,
    INRAE, 2020. Migale bioinformatics Facility, doi:
    10.15454/1.5572390655343293E12) for providing computing and storage resources.
    Louis Lacoste was funded by the MathNum department of INRAE and Fondation
    Mathématique Jacques Hadamard.

    \printbibliography
    \newpage
\fi

\ifNOSUPP
\else

\pagenumbering{arabic}
\renewcommand{\thepage}{S-\arabic{page}}
\renewcommand{\thefigure}{S-\arabic{figure}}
\renewcommand{\thetable}{S-\arabic{table}}
\renewcommand{\theequation}{S-\arabic{equation}}
\setcounter{figure}{0}
\setcounter{table}{0}
\setcounter{equation}{0}
\appendix
\section{Proof of the idenfiability result}
\label{sec:proof-of-the-idenfiability-result}

Along with results presented in~\Cref{prop:identifiability-iid-pirho} we also
have the identifiability results for $\pi$-colBiSBM and $\rho$-colBiSBM shown
in~\Cref{prop:identifiability-pi-rho}.

\begin{property}
    \label{prop:identifiability-pi-rho}

   {\normalfont $\pi$-colBiSBM} The parameters $\bm{\theta} = (\bm{\pi}^1, \dots, \bm{\pi}^M, \bm{\rho}, \bm{\alpha})$ are identifiable up to a
   label switching of the blocks if those conditions are achieved:
   \begin{enumerate}[label=(2.\arabic*)]
       \item $\forall m \in\{1,\dots,M\} : n^{m}_1 \geq 2 Q_2 - 1~\text{and}~\exists m^*\in\{1,\dots,M\} : n^{m^*}_2 \geq 2 Q_1 - 1$.\label{item:pi-n}
       \item The coordinates of the vector $\bm{\alpha}\bm{\rho}$ are distinct.\label{item:pi-tau1-complete}
       \item $\forall m \in \{1,\dots,M\},$ the coordinates of vector $(\overline{\bm{\pi}}^m)^{\top}\overline{\bm{\alpha}}^m$ are distinct.\label{item:pi-tau2}
       \item The coordinates of the vector $\bm{\rho}$ are distinct.\label{item:pi-order}
   \end{enumerate}%

   {\normalfont $\rho$-colBiSBM} The parameters $\bm{\theta} = (\bm{\pi}, \bm{\rho}^1, \dots, \bm{\rho}^M, \bm{\alpha})$ are identifiable up to a
   label switching of the blocks if those conditions are achieved:
    \begin{enumerate}[label=(3.\arabic*)]
       \item  $\exists m^{*} \in\{1,\dots,M\} : n^{m^{*}}_1 \geq 2 Q_2 - 1~\text{and}~\forall m\in\{1,\dots,M\} : n^{m}_2 \geq 2 Q_1 - 1$.\label{item:rho-n}
       \item $\forall m \in \{1,\dots,M\}$, the coordinates of vector $\overline{\bm{\alpha}}^m\overline{\bm{\rho}}^m$ are distinct.\label{item:rho-tau1}
       \item The coordinates of the vector $\bm{\pi}^{\top}\bm{\alpha}$ are distinct.\label{item:rho-tau2-complete}
       \item The coordinates of the vector $\bm{\pi}$ are distinct.\label{item:rho-order}
    \end{enumerate}%

\end{property}

\begin{proof}
    We derive the results \ifSUPPONLY of identifiability\else in~\Cref{prop:identifiability-iid-pirho,prop:identifiability-pi-rho}\fi.
    \paragraph{\normalfont iid-colBiSBM} \Cite{keribinEstimationSelectionLatent2015} building
    on~\cite{celisseConsistencyMaximumlikelihoodVariational2012}, proved that the
    parameters $(\bm{\pi}^m, \bm{\rho}^m, \bm{\alpha}^m)$ of the
    $\mathcal{F}\text{-biSBM}_{n_1^m,n_2^m}(Q_1^{m}, Q_2^{m}, \bm{\pi}^m, \bm{\rho}^m,
        \bm{\alpha}^m)$ are identifiable from the observation of network $X^m$ when
    $\mathcal{F}$ is the Bernoulli distribution and the following conditions are
    met:
    \begin{enumerate}
        \item $ n_1^m \geq 2 Q_2^{m} - 1~\text{and}~n_2^m \geq 2 Q_1^{m} - 1$.
        \item $\forall 1\leq q \leq Q_1^{m}, \pi_q^m > 0$
              and the coordinates of vector $
                  \bm{\alpha}^m\bm{\rho}^m$ are distinct.
        \item $\forall 1\leq r \leq Q_2^{m}, \rho_r^m > 0$
              and the coordinates of vector $(\bm{\pi}^m)^{\top}
                  \bm{\alpha}^m$ are distinct.
    \end{enumerate}

    Under the iid-colBiSBM model, for all $m=1\dots M$, $X^m \sim
        \mathcal{F}\text{-biSBM}_{n_1^m,n_2^m}(Q_1, Q_2, \bm{\pi}, \bm{\rho},
        \bm{\alpha})$. This means that
    applying~\cite{keribinEstimationSelectionLatent2015}, the identifiability of
    $\bm{\alpha}$, $\bm{\pi}$ and $\bm{\rho}$ is obtained from the distribution of
    $X^{m^*}$ under assumptions~\labelcref{item:iid-n,item:iid-tau1,item:iid-tau2}.

    \paragraph{\normalfont$\pi$-colBiSBM}
    Under $\pi$-colBiSBM, we have for all $m$,
    \[
        X^m \sim \mathcal{F}\text{-biSBM}_{n_1^m, n_2^m}(Q_1^{m}, Q_2, \overline{\bm{\pi}}^m, \bm{\rho}, \overline{\bm{\alpha}}^m)
    \]
    where $\overline{\bm{\pi}}^m$ is the vector of non-zero proportions of length
    $Q_1^{m}$ in $\bm{\pi}$ and $\overline{\bm{\alpha}}^m$ is the restriction of
    $\bm{\alpha}$ to the set $\mathcal{Q}_1^m$, i.e., the rows $(\alpha_{qr})_{1\le
                r\le Q_2}$ with $q \in \mathcal{Q}_1^m$. Under
    assumptions~\labelcref{item:pi-n,item:pi-tau1-complete,item:pi-tau2} and by
    noticing that~\labelcref{item:pi-tau1-complete} implies that $\forall m \in
        \{1,\dots,M\}$, the coordinates of vector $\overline{\bm{\alpha}}^m \bm{\rho}$
    are distinct, applying~\cite{keribinEstimationSelectionLatent2015} allows us to
    identify $\overline{\bm{\alpha}}^m$, $\overline{\bm{\pi}}^m$ and $\bm{\rho}$
    for each of the networks $X^m$. From the $\overline{\bm{\alpha}}^m$ we build
    $\bm{\alpha}$ the complete matrix that merges the $\overline{\bm{\alpha}}^m$.
    This is done by starting from any of the $\overline{\bm{\alpha}}^m$ and adding
    the missing entries from the others $\overline{\bm{\alpha}}^{m'}$ using
    assumption~\labelcref{item:pi-order} to reorder the columns and
    assumption~\labelcref{item:pi-tau1-complete} to reorder the rows. This way we
    build $\bm{\alpha}$ and have the index matches to define $\phi_m :
        \{1,\dots,Q_1^{m}\} \rightarrow \{1,\dots,Q_1\}$ such that
    $\overline{\alpha}_{\phi_m(q),r}^m = \alpha_{qr}$. Finally, we define
    $\bm{\pi}^m$ as follows:
    \[
        \pi^m_q = \begin{cases*}
            0                                 & $\forall q \in \{1,\dots,Q_1\} \setminus \phi_m(\{1,\dots,Q_1^{m}\})$ \\
            \overline{\pi}^m_{\phi_m^{-1}(q)} & $\forall q \in \phi_m(\{1,\dots,Q_1^{m}\})$
        \end{cases*}.
    \]

    \paragraph{\normalfont$\rho$-colBiSBM} Under $\rho$-colBiSBM, we have for all $m$,
    \[
        X^m \sim \mathcal{F}\text{-biSBM}_{n_1^m, n_2^m}(Q_1, Q_2^{m}, \bm{\pi}, \overline{\bm{\rho}}^m, \overline{\bm{\alpha}}^m)
    \]
    where $\overline{\bm{\rho}}^m$ is the vector of non-zero proportions of length
    $Q_2^{m}$ in $\bm{\rho}$ and $\overline{\bm{\alpha}}^m$ is the restriction of
    $\bm{\alpha}$ to the set $\mathcal{Q}_2^m$, i.e., the columns
    $(\alpha_{qr})_{1\le q\le Q_1}$ with $r\in\mathcal{Q}_2^m$. By
    applying~\cite{keribinEstimationSelectionLatent2015} using
    assumptions~\labelcref{item:rho-n,item:rho-tau1,item:rho-tau2-complete} and
    noticing that \labelcref{item:rho-tau2-complete} implies that $\forall m \in
        \{1,\dots,M\},$ the coordinates of vector
    $\bm{\pi}^{\top}\overline{\bm{\alpha}}^m$ are distinct, we identify
    $\overline{\bm{\alpha}}^m$, $\overline{\bm{\rho}}^m$ and $\bm{\pi}$ for each of
    the networks $X^m$. From the $\overline{\bm{\alpha}}^m$ we build $\bm{\alpha}$
    the complete matrix that merges the $\overline{\bm{\alpha}}^m$. This is done by
    starting from any of the $\overline{\bm{\alpha}}^m$ and adding the missing
    entries from the others $\overline{\bm{\alpha}}^{m'}$ using
    assumption~\labelcref{item:rho-order} to reorder the rows and
    assumption~\labelcref{item:rho-tau2-complete} to reorder the columns. This way
    we build $\bm{\alpha}$ and have the index matches to define $\psi_m :
        \{1,\dots,Q_2^{m}\} \rightarrow \{1,\dots,Q_2\}$ such that
    $\overline{\alpha}_{q\psi_m(r)}^m = \alpha_{qr}$. Finally, we define
    $\bm{\rho}^m$ as follows:
    \[
        \rho^m_r = \begin{cases*}
            0                                  & $\forall r \in \{1,\dots,Q_2\} \setminus \psi_m(\{1,\dots,Q_2^{m}\})$ \\
            \overline{\rho}^m_{\psi_m^{-1}(r)} & $\forall r \in \psi_m(\{1,\dots,Q_2^{m}\})$
        \end{cases*}.
    \]

    \paragraph{\normalfont$\pi\rho$-colBiSBM} Under $\pi\rho$-colBiSBM, we have for all $m$,
    \[
        X^m \sim \mathcal{F}\text{-biSBM}_{n_1^m, n_2^m}(Q_1^{m}, Q_2^{m}, \overline{\bm{\pi}}^m, \overline{\bm{\rho}}^m, \overline{\bm{\alpha}}^m)
    \]
    where $\overline{\bm{\pi}}^m$ is the vector of non-zero proportions of length
    $Q_1^{m}$ in $\bm{\pi}$, $\overline{\bm{\rho}}^m$ is the vector of non-zero
    proportions of length $Q_2^{m}$ in $\bm{\rho}$ and $\overline{\bm{\alpha}}^m$
    is the restriction of $\bm{\alpha}$ to the set
    $\mathcal{Q}_1^m\times\mathcal{Q}_2^m$, i.e,
    $(\alpha_{qr})_{q\in\mathcal{Q}_1^m,r\in\mathcal{Q}_2^m}$. By
    applying~\cite{keribinEstimationSelectionLatent2015} using
    assumptions~\labelcref{item:pirho-size,item:pirho-distinct-row,item:pirho-distinct-col}
    we have the identifiability of each of the $X^m$ networks. The difficulty lies
    in the reordering of the blocks across the networks and the fact that the
    $\overline{\bm{\alpha}}^m$ only represent subparts of the $\bm{\alpha}$ matrix.
    Under assumption~\ref{item:pirho-alpha} one can build $\bm{\alpha}$ from the
    $\overline{\bm{\alpha}}^m$.

    Start by taking $\overline{\bm{\alpha}}^1$ as the first $\bm{\alpha}$. Then for
    each $m\in\{2,\dots,M\}$, find common entries with the current $\bm{\alpha}$,
    from it deduce the permutations of rows and columns for
    $\overline{\bm{\alpha}}^m$ that will allow to match the entries. If there are
    new entries add them to the current $\bm{\alpha}$ either from the permutations,
    either at the end of the rows and/or columns. Update the previous index matches
    from previous networks that where possibly changed and repeat for next $m$. As
    mentioned\ifSUPPONLY~previously\else
    ~in~\Cref{ssec:a-collection-of-bipartite-sbm-with-varying-block-size-on-rows-or-columns}\fi~some
    row and column groups may never interact and thus $\bm{\alpha}$ would be
    incomplete.

    Building $\bm{\alpha}$ in this manner gives us the index matches and thus
    define $\phi_m : \{1,\dots,Q_1^{m}\} \rightarrow \{1,\dots,Q_1\}$ and $\psi_m
        : \{1,\dots,Q_2^{m}\} \rightarrow \{1,\dots,Q_2\}$ such that
    $\overline{\alpha}_{\phi_m(q)\psi_m(r)}^m = \alpha_{qr}$. Finally, we define
    $\bm{\pi}^m$ and $\bm{\rho}^m$ as follows:
    \begin{align*}
        \pi^m_q  & = \begin{cases*}
                         0                                 & $\forall q \in \{1,\dots,Q_1\} \setminus \phi_m(\{1,\dots,Q_1^{m}\})$ \\
                         \overline{\pi}^m_{\phi_m^{-1}(q)} & $\forall q \in \phi_m(\{1,\dots,Q_1^{m}\})$
                     \end{cases*} \quad \text{and}  \\
        \rho^m_r & = \begin{cases*}
                         0                                  & $\forall r \in \{1,\dots,Q_2\} \setminus \psi_m(\{1,\dots,Q_2^{m}\})$ \\
                         \overline{\rho}^m_{\psi_m^{-1}(r)} & $\forall r \in \psi_m(\{1,\dots,Q_2^{m}\})$
                     \end{cases*}.
    \end{align*}
\end{proof}

\clearpage
\section{On the model selection criterion}
\label{sec:on-the-model-selection-criterion}
For $\pi\rho$-colBiSBM, the model is described by its supports $\supp{1},
    \supp{2}$. Let us denote both supports by $\bothsupps$.

\newcommand{\eqpartalpha}{A}
\newcommand{\eqpartpi}{B1}
\newcommand{\eqpartrho}{B2}

\begin{equation}
    \begin{aligned}
        p(\bX, \bZ, \bW\mid\bothsupps)
         & = \int_{\thetasupp} p(\bX, \bZ, \bW\mid\bothsupps)\,p(\thetasupp) \,\mathrm{d}\thetasupp                                     \\
         & = \int_{\alphasupp}\!\int_{\pisupp}\!\int_{\rhosupp}
        p(\bX\mid\bZ,\bW,\bothsupps)\,p(\bZ\mid\pisupp,\supp{1})\,p(\bW\mid\rhosupp,\supp{2})                                           \\
         & \qquad\qquad\qquad\qquad\qquad\qquad
        p(\alphasupp)\,p(\pisupp)\,p(\rhosupp)\,\mathrm{d}\alphasupp\,\mathrm{d}\pisupp\,\mathrm{d}\rhosupp                             \\
         & = \underbrace{\int_{\alphasupp} p(\bX\mid\bZ,\bW,\alphasupp,\bothsupps)\,p(\alphasupp)\,\mathrm{d}\alphasupp}_{\eqpartalpha}
        \\
         & \qquad \times
        \underbrace{\int_{\pisupp} p(\bZ\mid\pisupp,\supp{1})\,\mathrm{d}\pisupp}_{\eqpartpi}
        \times\underbrace{\int_{\rhosupp} p(\bW\mid\rhosupp,\supp{2})\,\mathrm{d}\rhosupp}_{\eqpartrho},
    \end{aligned}
    \label{eqn:complete-likelihood-conditional-supports}
\end{equation}

We use a BIC approximation on~$\eqpartalpha$:
\begin{equation*}
    \begin{aligned}
        \eqpartalpha = p(\bX\mid\bZ,\bW,\bothsupps)
         & = \int_{\alphasupp}
        \Biggl( \prod_{m=1}^{M} p(X^m\mid Z^m, W^m; \alphasupp, \bothsupps) \Biggr)
        p(\alphasupp)\,\mathrm{d}\alphasupp                            \\
         & \approx \max_{\alphasupp} \exp\Biggl(
        \sum_{m=1}^{M} \ell(X^m \mid Z^m, W^m; \alphasupp, \bothsupps) \\
         & \qquad\qquad\qquad
        - \tfrac{1}{2} \underbrace{\sum_{q\in\mathcal{Q}_1}\sum_{r\in\mathcal{Q}_2}
        \mathbbb{1}_{((\supp{1})^{\prime}\supp{2})_{qr}>0}}_{\text{dim}(\alphasupp)} \,
        \log\Bigl(\sum_{m=1}^{M} n_1^m n_2^m\Bigr)
        + \mathcal{O}(1)
        \Biggr).
    \end{aligned}
\end{equation*}

For~$\eqpartpi$ and~$\eqpartrho$ we propose a $Q_{i}^m$-dimensional Dirichlet
prior for each mixture parameter:
\begin{equation*}
    \begin{aligned}
        \eqpartpi = p(\bZ \mid \supp{1})
         & = \prod_{m=1}^{M} \int_{\pi^m_{\supp{1}}} p(Z^m\mid\pi^m_{\supp{1}})\,p(\pi^m_{\supp{1}})\,\mathrm{d}\pi^m_{\supp{1}}     \\
         & \approx \max_{\pisupp} \exp\Biggl(
        \sum_{m=1}^{M} \ell(Z^m; \pisupp)
        - \tfrac{Q_1^{m} - 1}{2} \log(n_1^m)
        + \mathcal{O}(1)
        \Biggr),                                                                                                                     \\
        \eqpartrho = p(\bW \mid \supp{2})
         & = \prod_{m=1}^{M} \int_{\rho^m_{\supp{2}}} p(W^m\mid\rho^m_{\supp{2}})\,p(\rho^m_{\supp{2}})\,\mathrm{d}\rho^m_{\supp{2}} \\
         & \approx \max_{\rhosupp} \exp\Biggl(
        \sum_{m=1}^{M} \ell(W^m; \rhosupp)
        - \tfrac{Q_2^{m} - 1}{2} \log(n_2^m)
        + \mathcal{O}(1)
        \Biggr).
    \end{aligned}
\end{equation*}

Then, using the log of~\eqref{eqn:complete-likelihood-conditional-supports} and
inputting $\eqpartalpha,\eqpartpi$ and $\eqpartrho$:
\begin{equation}
    \begin{aligned}
        \log p(\bX,\bZ,\bW\mid\bothsupps)
         & \approx \max_{\alphasupp}\Biggl(
        \sum_{m=1}^{M} \ell(X^m \mid Z^m, W^m; \alphasupp, \bothsupps)                                                                 \\
         & \qquad\qquad\qquad\qquad
        - \tfrac{1}{2} \sum_{q,r=1}^{Q_1,Q_2}
        \mathbbb{1}_{((\supp{1})^{\prime}\supp{2})_{qr}>0} \,
        \log\Bigl(\sum_{m=1}^{M} n_1^m n_2^m\Bigr)
        \Biggr)                                                                                                                        \\
         & \quad + \max_{\pisupp}\Biggl(
        \sum_{m=1}^{M} \ell(Z^m; \pisupp) - \tfrac{Q_1^{m} - 1}{2} \log(n_1^m)
        \Biggr)                                                                                                                        \\
         & \quad + \max_{\rhosupp}\Biggl(
        \sum_{m=1}^{M} \ell(W^m; \rhosupp) - \tfrac{Q_2^{m} - 1}{2} \log(n_2^m)
        \Biggr)                                                                                                                        \\
         & \approx \max_{\thetasupp} \sum_{m=1}^{M}
        \Bigl[ \ell(X^m \mid Z^m, W^m,\bothsupps; \alphasupp)+ \ell(Z^m; \pisupp) + \ell(W^m; \rhosupp) \Bigr]                         \\
         & \qquad\qquad\qquad
        - \tfrac{1}{2}\Bigl(
        \sum_{m=1}^{M} (Q_1^{m} - 1) \log(n_1^m)
        + \sum_{m=1}^{M} (Q_2^{m} - 1) \log(n_2^m)                                                                                   \\
         & \qquad\qquad\qquad\qquad\qquad
        + \sum_{q= 1}^{Q_1}\sum_{r=1}^{Q_2}
        \mathbbb{1}_{((\supp{1})^{\prime}\supp{2})_{qr}>0} \,
        \log\Bigl(\sum_{m=1}^{M} n_1^m n_2^m\Bigr)
        \Bigr)                                                                                                                         \\
         & \approx \max_{\thetasupp} \ell(\bX,\bZ,\bW\mid\bothsupps;\thetasupp)                                                        \\
         & \qquad - \tfrac{1}{2} \bigl[\pen_{\pi}(Q_1,\supp{1}) + \pen_{\rho}(Q_2,\supp{2}) + \pen_{\alpha}(Q_1,Q_2,\bothsupps)\bigr].
    \end{aligned}
\end{equation}

Recalling that the penalties are:

\[ \text{pen}_{\pi}(Q_1, \supp{1}) = \sum_{m=1}^{M} (Q_{1}^{m} - 1)
    \log n_{1}^{m},
    ~\text{pen}_{\rho}(Q_2, \supp{2}) = \sum_{m=1}^{M} (Q_{2}^{m} - 1)
    \log n_{2}^{m},
\]
and
\[
    \text{pen}_{\alpha}(Q_1, Q_2, \bothsupps) = \left(\sum_{q=1}^{Q_1}
    \sum_{r=1}^{Q_2} \mathbbb{1}_{((\supp{1})'\supp{2})_{qr} > 0}\right) \log (N_M),
\]
with $N_M = \sum_{m = 1}^{M} n_{1}^{m} n_{2}^{m}$. As $\bZ$ and $\bW$ are
unknown we replace the $Z_{iq}^m, W_{jr}^m$ by their corresponding variational
parameters $\tau^{1,m}_{iq}, \tau^{2,m}_{jr}$:

\begin{equation*}
    \begin{aligned}
        \log p(\bX,\bZ,\bW\mid\bothsupps) & \approx \max_{\thetasupp} \E_{\Rest}[\ell(\bX,\bZ,\bW\mid\bothsupps;\thetasupp)]                                           \\
                                          & \qquad - \tfrac{1}{2} \bigl[\pen_{\pi}(Q_1,\supp{1}) + \pen_{\rho}(Q_2,\supp{2}) + \pen_{\alpha}(Q_1,Q_2,\bothsupps)\bigr] \\
                                          & \eqcolon \ICL(\bX,Q_1,Q_2,\bothsupps)
    \end{aligned}
\end{equation*}

To obtain the criterion $\BICL(\bX,Q_1,Q_2)$ we need to penalize for the
supports $\supp{1},\supp{2}$. For a given $Q_i$ blocks, where $i\in\{1,2\}$, we
set the prior on $\supp{i}$ to be the product of uniform priors for the number
of blocks $Q_i^{m}$ populated by the network $m$ and uniform priors for the
actual choice of the $Q_i^{m}$ blocks (i.e., the position of ones in the
$m$th row of $\supp{i}$) among the $Q_i$ possible blocks:

\begin{equation*}
    \begin{aligned}
        p_{Q_i}(\supp{i}) = p_{Q_i}(Q_1,\dots,Q_M) \cdot p_{Q_i}(\supp{i}\mid Q_1,\dots,Q_M) =  \frac{1}{Q_i^M} \prod_{m=1}^{M} \frac{1}{{Q_i \choose Q_i^{m}}}
    \end{aligned}
\end{equation*}

Using a BIC approximation and under a concentration assumption on the correct
supports we obtain:
\begin{equation*}
    \begin{aligned}
        \log p(\bX,\bZ,\bW) & = \log\int_{\supp{1}}\!\int_{\supp{2}} p(\bX,\bZ,\bW\mid\bothsupps)p_{Q_1}(\supp{1})p_{Q_2}(\supp{2})\di\supp{1}\di\supp{2}                                     \\
                            & \approx \log\int_{\supp{1}}\!\int_{\supp{2}}\exp\bigl[\ICL(\bX,Q_1,Q_2,\bothsupps)+\log(p_{Q_1}(\supp{1})) +\log(p_{Q_2}(\supp{2}))\bigr]\di\supp{1}\di\supp{2} \\
                            & \approx \max_{\supp{1}\in\admissupp{1},~\supp{2}\in\admissupp{2}} \bigl[\ICL(\bX,Q_1,Q_2,\bothsupps)+\log(p_{Q_1}(\supp{1}))+\log(p_{Q_2}(\supp{2}))\bigr].
    \end{aligned}
\end{equation*}

Recalling that:
\begin{equation*}
    \begin{aligned}
         & \ell(\bX,\bZ,\bW;\thetasupp) = \log p(\bZ,\bW\mid\bX;\thetasupp) + \ell(\bX;\thetasupp)                                         \\
         & \Rightarrow \E_{\bZ,\bW\mid\bX}[\ell(\bX,\bZ,\bW;\thetasupp)] = -\Entropy(p(\bZ,\bW\mid\bX;\thetasupp)) + \ell(\bX;\thetasupp),
    \end{aligned}
\end{equation*}
and under the variational approximation, namely approximating $\bZ,\bW\mid\bX$ by $\Rest$:
\begin{equation*}
    \begin{aligned}
        \E_{\Rest}[\ell(\bX,\bZ,\bW;\thetasupp)] = \ell(\bX;\thetasupp)-\Entropy(\Rest).
    \end{aligned}
\end{equation*}

Thus the ICL can be expressed as:
\begin{equation*}
    \begin{aligned}
        \ICL(\bX,Q_1,Q_2,\bothsupps) & = \max_{\thetasupp} \ell(\bX;\thetasupp)-\Entropy(\Rest)                                                                   \\
                                     & \qquad - \tfrac{1}{2} \bigl[\pen_{\pi}(Q_1,\supp{1}) + \pen_{\rho}(Q_2,\supp{2}) + \pen_{\alpha}(Q_1,Q_2,\bothsupps)\bigr] \\
    \end{aligned}
\end{equation*}

Finally in order to obtain the BIC-L criterion, we add the entropy of the
surrogate distribution as the ICL penalizes on the entropy and in order to
allow fuzzy node memberships we want to alleviate this penalization. By
denoting $\pen_{\supp{i}}(Q_i) = -2\log p_{Q_i}(\supp{i})$ and recalling that
$\vbound(\Rest, \thetasupp) = \E_{\Rest}[\ell(\bX,\bZ,\bW;\thetasupp)] +
    \Entropy(\Rest)$, the criterion is:
\begin{equation*}
    \begin{aligned}
        \BICL(\bX,Q_1,Q_2) & = \max_{\substack{\supp{1}\in\admissupp{1}                                                                         \\ \supp{2}\in\admissupp{2}}} \biggl[\max_{\thetasupp}\vbound(\Rest,\thetasupp) - \frac{1}{2}\bigl[\pen_{\pi}(Q_1,\supp{1}) + \pen_{\rho}(Q_2,\supp{2})\\
                           & \qquad\qquad\qquad\qquad\qquad\qquad\qquad\qquad\quad + \pen_{\alpha}(Q_1,Q_2,\bothsupps)                          \\
                           & \qquad\qquad\qquad\qquad\qquad\qquad\qquad\qquad\quad + \pen_{\supp{1}}(Q_1) + \pen_{\supp{2}}(Q_2) \bigr]\biggr].
    \end{aligned}
\end{equation*}

\section{Details of the partitioning algorithm}
\label{sec:details-of-the-network-partitioning-algorithm}

Continuing~\Cref{sssec:testing-common-connectivity-structure}, the principle is to adjust the colBiSBM model over the full collection of $M$
networks and then compute the dissimilarity matrix between all networks of the
collection. We obtain the collection $\mathcal{G} = \{1,\dots M\}$ the
trivial partition in a unique group.

The parameters for the dissimilarity are defined as follows:
\begin{align*}
    \widetilde{n}_{qr}^m & = \sum_{i=1}^{n_1^m} \sum_{j=1}^{n_2^m} \widehat{\tau}_{iq}^{1,m} \widehat{\tau}_{jr}^{2,m},
                         &                                                                                              & \widetilde{\alpha}_{qr}^m = \frac{\sum_{i=1}^{n_1^m} \sum_{j=1}^{n_2^m} \widehat{\tau}_{iq}^{1,m} \widehat{\tau}_{jr}^{2,m} X_{ij}^m}{\widetilde{n}_{qr}^m}, \\
    \widetilde{\pi}_q^m  & = \frac{\sum_{i=1}^{n_1^m} \widehat{\tau}_{iq}^{1,m}}{n_1^m},
                         &                                                                                              & \widetilde{\rho}_r^m = \frac{\sum_{j=1}^{n_2^m} \widehat{\tau_{jr}}^{2,m}}{n_2^m}.
\end{align*}
And the pairwise dissimilarity for networks $(m,m')\in\{1,\dots M\}^2$ is then:
\[
    D_{\mathcal{M}}(m,m') = \sum_{q = 1}^{Q_1} \sum_{r = 1}^{Q_2} \max(\widetilde{\pi}_{q}^{m}, \widetilde{\pi}_{q}^{m'}) \left( \widetilde{\alpha}_{qr}^{m} - \widetilde{\alpha}_{qr}^{m'}\right)^{2} \max(\widetilde{\rho}_{r}^{m}, \widetilde{\rho}_{r}^{m'})
\]

Then using a hierarchical clustering we split the collection in two
sub-collections with the dissimilarity matrix. On the two subcollections we
adjust a colBiSBM model and we compute the score of this new partition
$\mathcal{G}^{*} = \{G_1, G_2\}$. If $Sc(\mathcal{G}^{*}) > Sc(\mathcal{G})$,
we repeat the same procedure on $G_1$ and $G_2$. Else we return $\mathcal{G}$.
We illustrate our capacity to perform a partition of a simulated collection for
all colBiSBM models, results are presented
in~\cref{fig:ari-partitioning-boxplot} and illustrate that as soon as the
structure is sufficiently marked ($\eps[\alpha] \geq 0.3$) clustering is almost
perfect. For all details on the experiment
see~\Cref{ssec:clustering-of-simulated-networks}.

\section{Details of the numerical experiments}
\label{sec:details-of-numerical-experiments}

In this section, we conduct a comprehensive simulation study to evaluate the
accuracy of our inference procedure and to assess our ability to distinguish
between the various competing models. Additionally, we demonstrate the benefits
of jointly modeling multiple networks, particularly in terms of improved link
prediction performance.

\paragraph{Reproducibility} All the codes used to simulate and to perform the analysis can be found at
\url{https://github.com/Polarolouis/code-colbisbm/}.

\subsection{Assessing the quality of estimation}	\label{ssec:efficiency-of-the-inference}

We first assess our ability to fit the most complex model namely the
$\pi\rho$-colBiSBM in our framework. To do so, we simulate networks with
varying degrees of structural clarity and evaluate how well the model recovers
the underlying node clusters. Clustering is a central feature in network
analysis, and its quality is quantified using the Adjusted Rand Index (ARI). We
begin by describing the simulation settings, then introduce the quality
indicators, and finally present the results. Some tables are deferred to the
Supplementary Material section for clarity.

\paragraph{Simulation settings}
We set $M = 2$ and simulate bipartite networks involving $n_{1}^{m} = 240$
nodes in row and $n_{2}^{m} = 240$ nodes in column with $Q_1=4$ blocks in row
and $Q_2 = 4$ blocks in columns. The connection matrix $\bm{\alpha}$ is set as
follows:
\begin{align*}
     & \bm{\alpha} = .25 +
    \begin{pmatrix}
        3 \eps[\alpha] & 2 \eps[\alpha] & \eps[\alpha]   & - \eps[\alpha] \\
        2 \eps[\alpha] & 2 \eps[\alpha] & - \eps[\alpha] & \eps[\alpha]   \\
        \eps[\alpha]   & - \eps[\alpha] & \eps[\alpha]   & 2 \eps[\alpha] \\
        - \eps[\alpha] & \eps[\alpha]   & 2 \eps[\alpha] & 0
    \end{pmatrix},
\end{align*}
with $\eps[\alpha]$ taking nine equally spaced values ranging from 0 to 0.24. $\eps[\alpha]$ represents the strength of the network structure.If $\eps[\alpha] = 0$, the model is an Erdős–Rényi model with a constant probability of connexion between any pair of nodes.  As $\eps[\alpha]$ increases, the differences between the blocks become more pronounced, leading to a clearer underlying structure in the network.

For each unique combination of parameters, we perform $N=3$ repetitions under
the model $\mathcal{B}$ern-$BiSBM_{240,240}( 4, 4, \bm{\alpha}, \bm{\pi}^m,
    \bm{\rho}^m)$ where the block proportions $(\bm{\pi}^m)_{m=1,2}$ and
$(\bm{\rho}^m )_{m=1,2}$ are chosen as follows \begin{align*}
    \bm{\pi}^1 = \sigma_1
    \begin{pmatrix}
        0.2 & 0.4 & 0.4 & 0
    \end{pmatrix},
                               &  & \bm{\pi}^2 =
    \begin{pmatrix}
        0.25 & 0.25 & 0.25 & 0.25
    \end{pmatrix},                    \\
    \bm{\rho}^1 =
    \begin{pmatrix}
        0.25 & 0.25 & 0.25 & 0.25
    \end{pmatrix}, &  &
    \bm{\rho}^2 = \sigma_2
    \begin{pmatrix}
        0 & 0.33 & 0.33 & 0.33
    \end{pmatrix},    &  &
\end{align*}
where $\sigma_1$ and $\sigma_2$ are two permutations of $\{ 1, \dots , 4 \}$ randomly chosen for each one of the 3 repetitions.   This allows to randomly affect the non-present block in each simulation.


\paragraph{Inference}
For each network, we put in competition the five models iid-colBiSBM,
$\pi$-colBiSBM, $\rho$-colBiSBM, $\pi\rho$-colBiSBM $sep\text{-}biSBM$. The
parameters are estimated with the VEM algorithm described
in~\Cref{ssec:variational-estimation-of-the-parameters}. The numbers of blocks
$Q_1$ and $Q_2$ and if necessary the supports $S^1$ and $S^2$ are selected with
the BIC-L criterion. The BIC-L is also used to chose among the five models.


\paragraph{Results}
\Cref{fig:inference-prop-modele-pref} proves that the $\pi\rho$-colBiSBM
is the preferred model if the networks' block structure is clear enough. In  case where $\eps[\alpha]$ is very small, all the blocks are equivalent and
the best model is the iid-colBiSBM.

\Cref{tab:Sim1:selectQ} provides the frequencies of choice of the right number
of blocks, i.e., $\widehat{Q}_1 = \widehat{Q}_2=4$. As one can see, the structure is
not recovered for small values of $\eps[\alpha]$, which is not surprising since
in this case, the network is close to an Erdős-Rényi model. However, as
soon as $\eps[\alpha]$ is over $0.06$ the methods recovers the true numbers of blocks in
most cases (frequency higher than $0.61$). Even more remarkably, the $4-4$
block structure is also found in the majority of cases, even if the assumption
on the proportions is not the right one (columns 1, 2, and 3 of the table).

To go further, we evaluate the quality of the clustering of the nodes into the
blocks depending on the chosen model. To compare the inferred clustering
(denote $\widehat{\bm{Z}}$) with the simulated/true ones (denote $\bm{Z}^*$) ,
we use the Adjusted Rand Index (ARI,~\cite{hubertComparingPartitions1985}) which
is a measure of the similarity between two clusterings: $ARI = 0$ for a random
clustering while $ARI = 1$ for a perfect match between
clusterings\footnote{Note that even if Rand Index can only yield values between
    0 and 1, ARI can return negative values if the RI is less than the expected
    value. This indicates a structure in clustering discordance.}. For each
network, we compare the block memberships obtained respectively by the
$\pi$-colBiSBM, $\rho$-colBiSBM and $\pi\rho$-colBiSBM models to the true ones.
To do so, we compute the mean of the ARI per dimension (rows and cols) over the
$2$ networks:
\begin{eqnarray*}
    \overline{\text{ARI}}_{\text{row}} & = \frac{1}{2}\left[\text{ARI}(\widehat{\bm{Z}^1 },\bm{Z}^{*1} ) + \text{ARI}(\widehat{\bm{Z}^2 },\bm{Z}^{*2} ) \right],    \\
    \overline{\text{ARI}}_{\text{col}} & = \frac{1}{2} \left[  \text{ARI}(\widehat{\bm{W}^1 },\bm{W}^{*1} ) + \text{ARI}(\widehat{\bm{W}^2 },\bm{W}^{*2} ) \right].
\end{eqnarray*}
The evolutions of $\overline{\text{ARI}}_{row}$ and $\overline{\text{ARI}}_{col}$ as $\eps[\alpha]$ increases  are  plotted on the top line of \Cref{fig:inference-ari-plots}.
As we see, the intra network clustering of the nodes is very good for every model as soon as $\eps[\alpha]$ exceeds $0.09$. However, we have to check if the blocks are coherent from one network to another: i.e., we want to be sure the  block $k$ of network $1$ indeed corresponds to block $k$ in network $2$.
To do so, we concatenate the row node clustering $(\bm{Z}^1,\bm{Z}^2)$ and
compute the global ARI (across networks) in row and col as:
\begin{eqnarray*}
    \text{ARI}_{\text{row}} & =& \text{ARI}\big((\widehat{\bm{Z}^1},\widehat{\bm{Z}^2}),(\bm{Z}^{*1},\bm{Z}^{*2}) \big) \\
    \text{ARI}_{\text{col}} & =& \text{ARI}\big((\widehat{\bm{W}^1},\widehat{\bm{W}^2}),(\bm{W}^{*1},\bm{W}^{*2}) \big)
\end{eqnarray*}
The evolutions of $\text{ARI}_{row}$ and $\text{ARI}_{col}$ as $\eps[\alpha]$ increases  are  plotted in the bottom line of~\Cref{fig:inference-ari-plots}.
As expected, the global ARI of the sep-colBiSBM is around $\frac{1}{2}$. Indeed, in this case, no coherence between the networks is introduced and consequently,  due to label switching, the blocks are randomly matched leading to the observed phenomena.
When $\eps[\alpha]$ increases the global ARI for iid-colBiSBM and $\pi\rho$-colBiSBM stabilizes around $0.9$: the proposed colBiSBM are able to recover the common structure in the $2$ networks and gather nodes from various networks playing the same role. For $\pi$-colBiSBM and $\rho$-colBiSBM the stabilization occurs around 0.7 and 0.8, the differences with the best models can be interpreted as the need of either having no flexibility or complete flexibility on both axis to not hinder performances by introducing some possible confusion.

%
%
%
%
%
\begin{figure}[ht]
    \centering
    \includegraphics{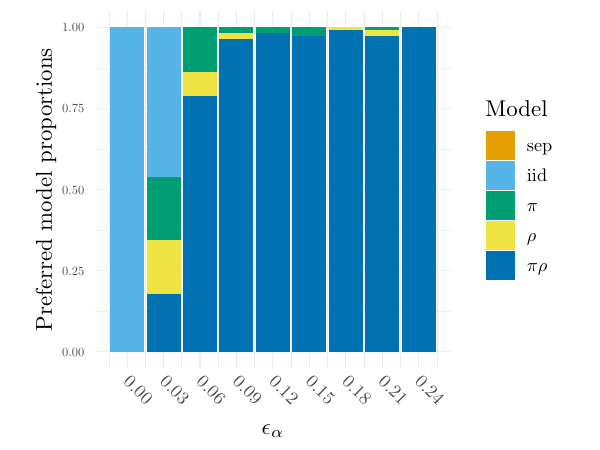}
    \caption{\textit{Assessing the quality of estimation}: Preferred model proportions over all datasets in function of
        $\eps[\alpha]$}
    \label{fig:inference-prop-modele-pref}
\end{figure}

\begin{figure}[htb]
    \centering
    \includegraphics[width=0.8\textwidth]{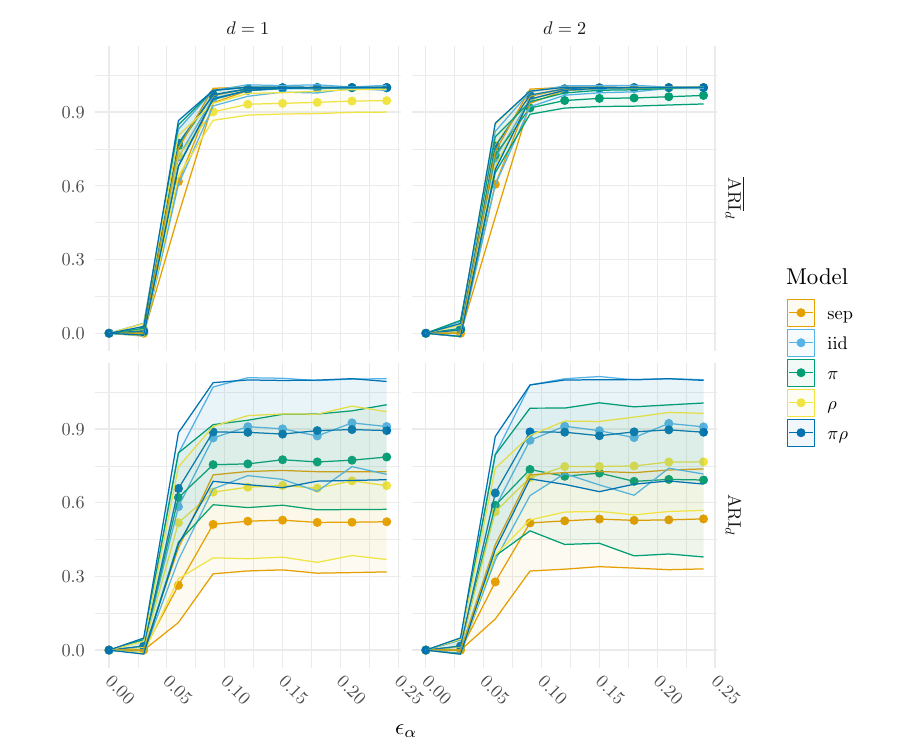}
    \caption{\textit{Assessing the quality of estimation}: Plot of the ARI quality indicators in function of
        $\eps[\alpha]$}
    \label{fig:inference-ari-plots}
\end{figure}

\begin{table}
\centering
\caption{\textit{Assessing the quality of estimation}: proportion of datasets where the correct numbers of blocks are selected.}
\centering
\fontsize{10}{12}\selectfont
\begin{tabular}[t]{|l|cc|cc|cc|cc|cccc|cccc|||l|cc|cc|cc|cc|cccc|cccc|||l|cc|cc|cc|cc|cccc|cccc|||l|cc|cc|cc|cc|cccc|cccc|||l|cc|cc|cc|cc|cccc|cccc|||l|cc|cc|cc|cc|cccc|cccc|||l|cc|cc|cc|cc|cccc|cccc|||l|cc|cc|cc|cc|cccc|cccc|||l|cc|cc|cc|cc|cccc|cccc|}
\hline
\multicolumn{1}{|c|}{ } & \multicolumn{2}{c|}{iid-colBiSBM} & \multicolumn{2}{c|}{$\pi$-colBiSBM} & \multicolumn{2}{c|}{$\rho$-colBiSBM} & \multicolumn{2}{c|}{$\pi\rho$-colBiSBM} \\
\cline{2-3} \cline{4-5} \cline{6-7} \cline{8-9}
$\epsilon_{\alpha}$ & $\bm{1}_{\widehat{Q_1} = 4}$ & $\bm{1}_{\widehat{Q_2} = 4}$ & $\bm{1}_{\widehat{Q_1} = 4}$ & $\bm{1}_{\widehat{Q_2} = 4}$ & $\bm{1}_{\widehat{Q_1} = 4}$ & $\bm{1}_{\widehat{Q_2} = 4}$ & $\bm{1}_{\widehat{Q_1} = 4}$ & $\bm{1}_{\widehat{Q_2} = 4}$\\
\hline
0.00 & 0 & 0 & 0 & 0 & 0 & 0 & 0 & 0\\
0.03 & 0 & 0 & 0 & 0 & 0 & 0 & 0 & 0\\
0.06 & 0.64 $\pm$ 0.05 & 0.61 $\pm$ 0.05 & 0.67 $\pm$ 0.05 & 0.64 $\pm$ 0.05 & 0.66 $\pm$ 0.05 & 0.59 $\pm$ 0.05 & 0.86 $\pm$ 0.03 & 0.86 $\pm$ 0.03\\
0.09 & 0.97 $\pm$ 0.02 & 0.94 $\pm$ 0.02 & 0.71 $\pm$ 0.04 & 1 & 1 & 0.64 $\pm$ 0.05 & 0.83 $\pm$ 0.04 & 0.88 $\pm$ 0.03\\
0.12 & 0.91 $\pm$ 0.03 & 0.91 $\pm$ 0.03 & 0.74 $\pm$ 0.04 & 1 & 1 & 0.66 $\pm$ 0.05 & 0.92 $\pm$ 0.03 & 0.91 $\pm$ 0.03\\
0.15 & 0.91 $\pm$ 0.03 & 0.94 $\pm$ 0.02 & 0.83 $\pm$ 0.04 & 1 & 1 & 0.71 $\pm$ 0.04 & 0.92 $\pm$ 0.03 & 0.94 $\pm$ 0.02\\
0.18 & 0.93 $\pm$ 0.03 & 0.89 $\pm$ 0.03 & 0.81 $\pm$ 0.04 & 1 & 1 & 0.73 $\pm$ 0.04 & 0.94 $\pm$ 0.02 & 0.94 $\pm$ 0.02\\
0.21 & 0.91 $\pm$ 0.03 & 0.92 $\pm$ 0.03 & 0.83 $\pm$ 0.04 & 1 & 1 & 0.7 $\pm$ 0.04 & 0.88 $\pm$ 0.03 & 0.9 $\pm$ 0.03\\
0.24 & 0.9 $\pm$ 0.03 & 0.92 $\pm$ 0.03 & 0.81 $\pm$ 0.04 & 1 & 1 & 0.79 $\pm$ 0.04 & 0.93 $\pm$ 0.03 & 0.93 $\pm$ 0.03\\
\hline
\end{tabular} \label{tab:Sim1:selectQ}
\end{table}

\subsection{From fixed  to varying block proportions }
\label{ssec:from-fixed-block-proportions-to-varying-block-proportions}
In this numerical experiment, we test our capacity to distinguish
$\pi\rho$-colBiSBM~ from iid-colBiSBM and other models
\paragraph{Simulation settings} To do so, we simulate $M=3$ networks of the same size involving $n_{1}^{m} =
    90$ row nodes and $n_{2}^{m} = 90$ column nodes. We fix a connection structure 
\begin{align*}
    \bm{\alpha} = \begin{pmatrix}
                      0.73 & 0.57 & 0.41 \\
                      0.57 & 0.57 & 0.09 \\
                      0.41 & 0.09 & 0.41
                  \end{pmatrix}.
\end{align*}
We vary the difficulty to decipher between the models by setting the block proportions as follows:
\begin{align*}
    \bm{\pi}^1 & = \begin{pmatrix}
                       \frac{1}{3}, & \frac{1}{3}, & \frac{1}{3}
                   \end{pmatrix},                         & \bm{\rho}^1 & = \begin{pmatrix}
                                                                                \frac{1}{3}, & \frac{1}{3}, & \frac{1}{3}
                                                                            \end{pmatrix},               \\
    \bm{\pi}^2 & = \sigma\begin{pmatrix}
                             \frac{1}{3} - \eps[\pi], & \frac{1}{3}, & \frac{1}{3} + \eps[\pi]
                         \end{pmatrix}, & \bm{\rho}^2 & = \sigma\begin{pmatrix}
                                                                    \frac{1}{3} - \eps[\rho], & \frac{1}{3}, & \frac{1}{3} + \eps[\rho]
                                                                \end{pmatrix}, \\
    \bm{\pi}^3 & = \sigma\begin{pmatrix}
                             \frac{1}{3} + \eps[\pi], & \frac{1}{3}, & \frac{1}{3} - \eps[\pi]
                         \end{pmatrix}, & \bm{\rho}^3 & = \sigma\begin{pmatrix}
                                                                    \frac{1}{3} + \eps[\rho] & \frac{1}{3}, & \frac{1}{3} - \eps[\rho]
                                                                \end{pmatrix},
\end{align*}
where $\eps[\pi]$ and $\eps[\rho]$ takes 9 values equally spaced in $\left[ 0,
        0.28\right]$.
For each combination of parameters, $N=3$ repetitions are conducted. Note that
when $\eps[\pi] = 0$ and $\eps[\rho] = 0$, the true model is the
iid-colBiSBM. When $\eps[\pi] > 0$ and $\eps[\rho] = 0$ the true model is a
$\pi$-colBiSBM. When $\eps[\pi] = 0$ and $\eps[\rho] > 0$ the true model is a
$\rho$-colBiSBM. Finally, when $\eps[\pi] > 0$ and $\eps[\rho] > 0$, the true
model is a $\pi\rho$-colBiSBM. The five models $\pi\rho$-colBiSBM,
$\pi$-colBiSBM, $\rho$-colBiSBM, iid-colBiSBM and $sep$-colBiSBM are fitted,
and the preferred model is chosen by the BIC-L criterion.

\begin{figure}[!hbt]
    \centering
    \includegraphics[width=0.85\textwidth]{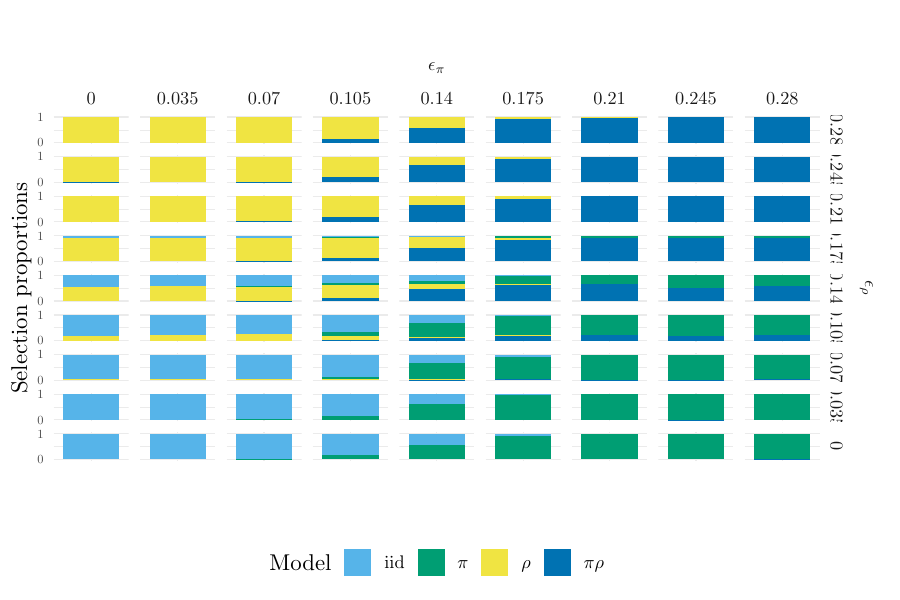}
    \caption{\label{fig:pref_model_func_eps}\textit{From fixed  to varying block proportions}: Plot of model selection proportions	over the different datasets in function of $\eps[\pi]$ and $\eps[\rho]$}
\end{figure}

\paragraph{Results}
\Cref{fig:pref_model_func_eps} shows the selection frequency of each
model across simulations. As expected, the sep-colBiSBM is never selected. When
$\eps[\pi]$ or $\eps[\rho]$ is close to 0, the iid-colBiSBM model is
predominantly selected. A noticeable transition occurs around $\eps[\pi] =
    0.14$, marking a shift in the preferred models.
Before this threshold, the iid-colBiSBM and the $\rho$-colBiSBM are frequently selected. Beyond this point, the $\pi$-colBiSBM
and $\pi\rho$-colBiSBM are increasingly favored.
A similar result is observed on the other axis around $\eps[\rho] = 0.14$
favoring the $\rho$-colBiSBM and $\pi\rho$-colBiSBM after this threshold.
Eventually, when the group proportions differ significantly across networks,
the $\pi\rho$-colBiSBM is systematically selected. The fact that both
$\eps[\pi]$ and $\eps[\rho]$ need to exceed 0.17 before the $\pi\rho$ model is
preferred suggests that a strong inter-network difference is required to
justify selecting this model.
The below~\Cref{tab:model-selection-block-recovery}
shows the retrieval percentage of the correct number of blocks in the model selection
experiment presented in~\Cref{ssec:model-selection}.

\begin{table}[!ht]
\centering
\caption{\textit{From fixed  to varying block proportions}. Proportion of dataset where the correct number of blocks is
        selected. Only shows results for $\epsilon_{\rho}\in \{0, 0.14, 0.28\}$.\label{tab:model-selection-block-recovery}}
\centering
\fontsize{9}{11}\selectfont
\begin{tabular}[t]{|ll|cc|cc|cc|cc|c|||ll|cc|cc|cc|cc|c|||ll|cc|cc|cc|cc|c|||ll|cc|cc|cc|cc|c|||ll|cc|cc|cc|cc|c|||ll|cc|cc|cc|cc|c|||ll|cc|cc|cc|cc|c|||ll|cc|cc|cc|cc|c|||ll|cc|cc|cc|cc|c|||ll|cc|cc|cc|cc|c|}
\hline
\multicolumn{2}{|c|}{ } & \multicolumn{2}{c|}{iid} & \multicolumn{2}{c|}{$\pi$} & \multicolumn{2}{c|}{$\rho$} & \multicolumn{2}{c|}{$\pi\rho$} \\
\cline{3-4} \cline{5-6} \cline{7-8} \cline{9-10}
$\epsilon_{\pi}$ & $\epsilon_{\rho}$ & $\bm{1}_{\widehat{Q_1}_{iid}=3}$ & $\bm{1}_{\widehat{Q_2}_{iid}=3}$ & $\bm{1}_{\widehat{Q_1}_{\pi}=3}$ & $\bm{1}_{\widehat{Q_2}_{\pi}=3}$ & $\bm{1}_{\widehat{Q_1}_{\rho}=3}$ & $\bm{1}_{\widehat{Q_2}_{\rho}=3}$ & $\bm{1}_{\widehat{Q_1}_{\pi\rho}=3}$ & $\bm{1}_{\widehat{Q_2}_{\pi\rho}=3}$\\
\hline
0.000 & 0.00 & 1 & 1 & 1 & 1 & 1 & 1 & 1 & 1\\
0.000 & 0.14 & 0.99 $\pm$ 0.01 & 1 & 1 & 1 & 1 & 1 & 1 & 1\\
0.000 & 0.28 & 0.93 $\pm$ 0.03 & 0.99 $\pm$ 0.01 & 0.81 $\pm$ 0.04 & 1 & 1 & 1 & 1 & 1\\
0.035 & 0.00 & 1 & 1 & 1 & 1 & 1 & 1 & 1 & 1\\
0.035 & 0.14 & 1 & 1 & 1 & 1 & 1 & 1 & 1 & 1\\
0.035 & 0.28 & 0.97 $\pm$ 0.02 & 0.96 $\pm$ 0.02 & 0.86 $\pm$ 0.03 & 1 & 1 & 1 & 1 & 1\\
0.070 & 0.00 & 1 & 1 & 1 & 1 & 1 & 1 & 1 & 1\\
0.070 & 0.14 & 1 & 1 & 1 & 1 & 1 & 1 & 0.99 $\pm$ 0.01 & 1\\
0.070 & 0.28 & 0.95 $\pm$ 0.02 & 0.96 $\pm$ 0.02 & 0.78 $\pm$ 0.04 & 1 & 1 & 1 & 0.99 $\pm$ 0.01 & 0.99 $\pm$ 0.01\\
0.105 & 0.00 & 1 & 1 & 1 & 1 & 1 & 1 & 1 & 1\\
0.105 & 0.14 & 1 & 1 & 1 & 1 & 1 & 1 & 1 & 1\\
0.105 & 0.28 & 0.96 $\pm$ 0.02 & 0.95 $\pm$ 0.02 & 0.79 $\pm$ 0.04 & 1 & 1 & 1 & 0.99 $\pm$ 0.01 & 0.99 $\pm$ 0.01\\
0.140 & 0.00 & 1 & 1 & 1 & 1 & 1 & 1 & 1 & 1\\
0.140 & 0.14 & 0.98 $\pm$ 0.01 & 0.99 $\pm$ 0.01 & 1 & 1 & 1 & 1 & 1 & 1\\
0.140 & 0.28 & 0.98 $\pm$ 0.01 & 0.94 $\pm$ 0.02 & 0.81 $\pm$ 0.04 & 1 & 1 & 0.99 $\pm$ 0.01 & 0.99 $\pm$ 0.01 & 1\\
0.175 & 0.00 & 1 & 1 & 1 & 1 & 1 & 1 & 1 & 1\\
0.175 & 0.14 & 0.98 $\pm$ 0.01 & 0.98 $\pm$ 0.01 & 0.99 $\pm$ 0.01 & 1 & 1 & 0.99 $\pm$ 0.01 & 1 & 1\\
0.175 & 0.28 & 0.98 $\pm$ 0.01 & 0.97 $\pm$ 0.02 & 0.82 $\pm$ 0.04 & 1 & 1 & 1 & 1 & 1\\
0.210 & 0.00 & 0.99 $\pm$ 0.01 & 0.98 $\pm$ 0.01 & 1 & 1 & 1 & 0.99 $\pm$ 0.01 & 1 & 1\\
0.210 & 0.14 & 0.99 $\pm$ 0.01 & 0.96 $\pm$ 0.02 & 0.99 $\pm$ 0.01 & 1 & 1 & 0.99 $\pm$ 0.01 & 1 & 1\\
0.210 & 0.28 & 0.94 $\pm$ 0.02 & 0.95 $\pm$ 0.02 & 0.75 $\pm$ 0.04 & 1 & 1 & 0.98 $\pm$ 0.01 & 0.99 $\pm$ 0.01 & 0.99 $\pm$ 0.01\\
0.245 & 0.00 & 0.99 $\pm$ 0.01 & 0.97 $\pm$ 0.02 & 1 & 1 & 1 & 0.97 $\pm$ 0.02 & 1 & 1\\
0.245 & 0.14 & 0.94 $\pm$ 0.02 & 0.95 $\pm$ 0.02 & 0.99 $\pm$ 0.01 & 1 & 1 & 0.98 $\pm$ 0.01 & 1 & 1\\
0.245 & 0.28 & 0.94 $\pm$ 0.02 & 0.93 $\pm$ 0.03 & 0.72 $\pm$ 0.04 & 1 & 1 & 0.94 $\pm$ 0.02 & 1 & 1\\
0.280 & 0.00 & 0.98 $\pm$ 0.01 & 0.92 $\pm$ 0.03 & 0.99 $\pm$ 0.01 & 1 & 1 & 0.86 $\pm$ 0.03 & 1 & 1\\
0.280 & 0.14 & 0.96 $\pm$ 0.02 & 0.95 $\pm$ 0.02 & 0.98 $\pm$ 0.01 & 1 & 1 & 0.86 $\pm$ 0.03 & 1 & 1\\
0.280 & 0.28 & 0.94 $\pm$ 0.02 & 0.96 $\pm$ 0.02 & 0.77 $\pm$ 0.04 & 1 & 1 & 0.73 $\pm$ 0.04 & 0.99 $\pm$ 0.01 & 0.98 $\pm$ 0.01\\
\hline
\end{tabular}
\end{table}

The proportions of correct block number selected stay relatively high for all
models and $\eps[\pi], \eps[\rho]$ but something interesting can be seen for
the $\pi$-colBiSBM and $\rho$-colBiSBM. When the other axis e.g., the columns
for the $\pi$-colBiSBM sees the proportions differences increased it diminishes
the correct retrieval rate for the free axis. One may suspect other causes but
the performance of the $\pi\rho$-colBiSBM being near perfect invites us to
point in the direction of a confusion induced in the free axis by the non-free
one.

\subsection{Information transfer between networks}
\label{ssec:information-transfer-between-networks}

One of the motivations for modeling collections of networks jointly rather
than separately is the potential for \textit{information transfer} between them.
To illustrate this advantage, we evaluate how information can be transferred
from a large network to a smaller one with missing values, aiming to recover
the fine structure of the smaller network. In particular, we compare the
performance of different methods in retrieving the node blocks affected by
missing edges.

\subsubsection{On modular structure}\label{sssec:on-modular-structure}
\paragraph{Simulation settings}

For this purpose we generate collections with two networks ($M=2$) of size
$n^{1}_1 = n^{1}_2 = 20$ and $n^{2}_1 = n^{2}_2 = 120$. One collection
is generated for each colBiSBM model. For each colBiSBM model, the experiment is
repeated $N=10$ times.

\begin{align*}
    \bm{\pi}^m = \begin{cases}
                     \bm{\pi} = \left( 0.5, 0.3, 0.2 \right) & \text{for iid}                     \\
                     \sigma_1^m(\bm{\pi})                    & \text{for } \pi \text{ and } \pi\rho
                 \end{cases}, &\qquad
    \bm{\rho}^m =
    \begin{cases}
        \bm{\rho}  = \left( 0.5, 0.3, 0.2 \right) & \text{for iid}                       \\
        \sigma_2^m(\bm{\rho})                     & \text{for } \rho \text{ and } \pi\rho
    \end{cases},
\end{align*}
for the block proportions, and two different structures with the corresponding
$\bm{\alpha}$,
\[
    \bm{\alpha}^{modular} = \begin{pmatrix}
        0.9  & 0.05 & 0.05 \\
        0.05 & 0.2  & 0.05 \\
        0.05 & 0.05 & 0.8
    \end{pmatrix}
\]
where $\bm{\alpha}^{modular}$ represents networks with community structure,
i.e., pairs of blocks that tend to interact preferentially within the block and
less outside of the blocks. In network $m=1$ (i.e.,
the smallest one) a proportion of the edges $p_{\text{mis}}$ see their values
replaced by \texttt{NA}s. 

\paragraph{Test procedure} A standard biSBM is fitted on the first network alone, and the predicted block
memberships are saved, along with the predicted links using the inferred
parameters. This will serve as a baseline to see if the use of the collection
benefits the predictions. A colBiSBM model is then fitted of the two networks
and we store the then obtained predictions of the missing links of network
$m=1$.

\paragraph{Quality metrics} To benchmark the performance we use the \emph{ROC Area Under the Curve} (ROC
AUC) for predicted versus real link values and the ARI that compare the
predicted block memberships to the true block memberships.

\begin{figure}[htb]
    \centering
    \includegraphics{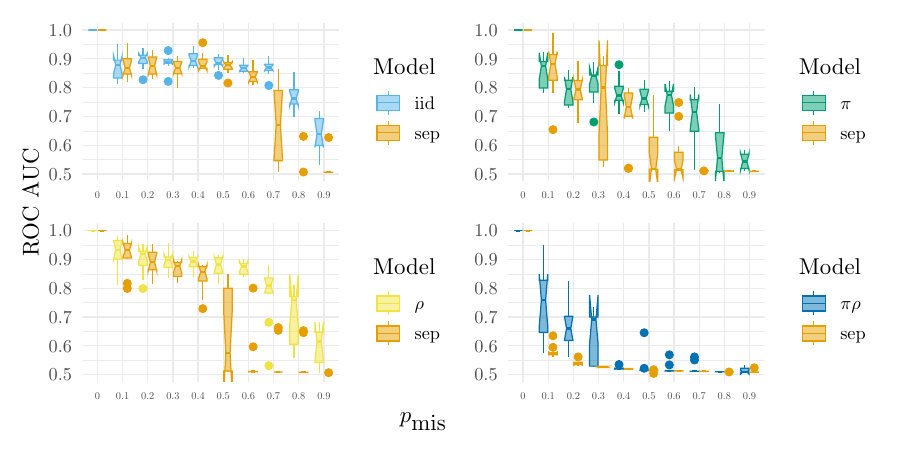}
    \caption{\textit{Information transfer on modular structure}: ARI over rows and columns in function of $\pmis$, the proportion of missing links
        for the four colBiSBM models and their LBM counterparts.
        Each dataset is generated with the corresponding colBiSBM model on a modular structure and
        fitted with the same model and the LBM model labelled as sep-colBiSBM to give a baseline of the performance.
    }%
    \label{fig:auc-plot-modular}
\end{figure}

\begin{figure}[H]
    \centering
    \includegraphics{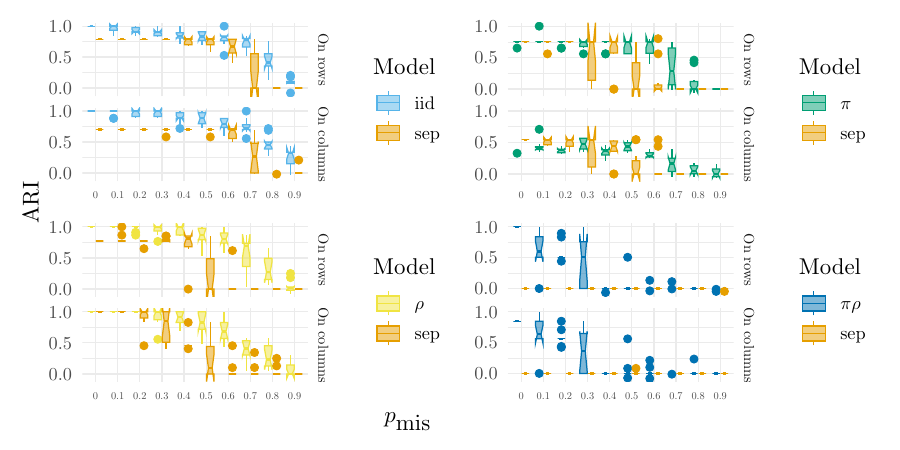}
    \caption{\textit{Information transfer on modular structure}: AUC over rows and columns in function of $\pmis$, the proportion of missing links
        for the four colBiSBM models and their LBM counterparts.
        Each dataset is generated with the corresponding colBiSBM model on a modular structure and
        fitted with the same model and the LBM model labelled as sep-colBiSBM to give a baseline of the performance.
    }%
    \label{fig:ari-plot-modular}
\end{figure}

\paragraph{Results}
\Cref{fig:auc-plot-modular,fig:ari-plot-modular}
show the results for the colBiSBM model and its sep-biSBM counterpart  fitted on
the same data.

For the AUC~(\Cref{fig:auc-plot-modular}), in almost all cases and for almost
all models the differences are not significant, but our models seem to perform
marginally better than their LBM counterpart and with a smaller variance. The
colBiSBM models seem able to endure higher degradation levels than the sep
counterpart. This
indicates that information is transferred from the bigger network that can
learn the full structure when estimating the parameters and predicting link
values.

For the ARI~(\Cref{fig:ari-plot-modular}), our models almost always do at least
as good as the sep counterpart and with a smaller variance. The block
membership attribution for rows for $\pi$-colBiSBM reveals more challenging and
yields an ARI of at most 0.75. And for the $\pi\rho$ datasets the sep is
completely random ($\text{ARI} = 0$) and the $\pi\rho$-colBiSBM is only better
for the very low proportions of missing data.

\subsubsection{Simulation on nested structure}\label{sssec:simulation-on-nested-structure}

\paragraph{Simulation settings} We use the same simulation settings, test procedure and quality metrics as
in~\Cref{ssec:information-transfer-between-networks}. The only difference being
the connectivity structure,
\[
    \bm{\alpha}^{nested} = \begin{pmatrix}
                               0.9  & 0.65 & 0.1  \\
                               0.35 & 0.15 & 0.05 \\
                               0.1  & 0.05 & 0.05
                           \end{pmatrix}
\]
where $\bm{\alpha}^{nested}$ represents a common structure detected in ecology
with generalist and specialist species and a \enquote{nested} structure.

\begin{figure}[!bt]
    \centering
    \includegraphics{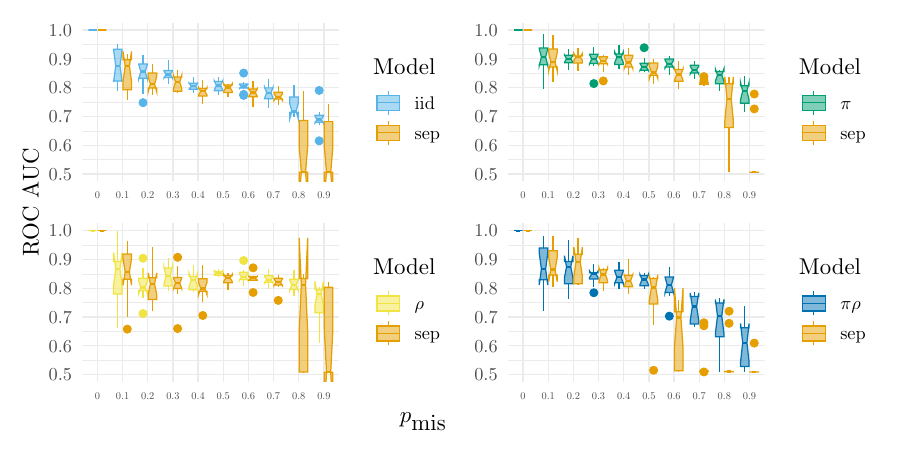}
    \caption{\textit{Information transfer on nested structure}: ROC AUC in function of $\pmis$, the proportion of missing links
        for the four colBiSBM models and their LBM counterparts.
        Each dataset is generated with the corresponding colBiSBM model on a nested structure and
        fitted with the same model and the LBM model labelled as sep to give a baseline of the performance.}%
    \label{fig:auc-plot-nested}
\end{figure}

\begin{figure}[!bt]
    \centering
    \includegraphics{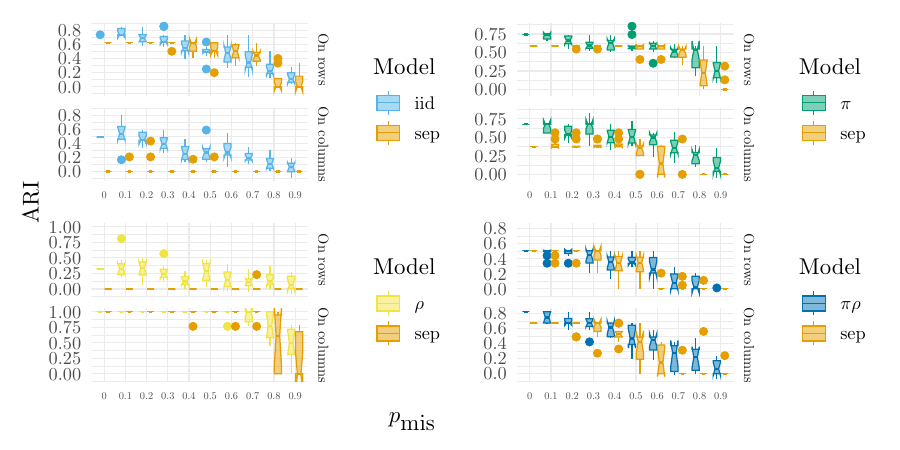}
    \caption{\textit{Information transfer on nested structure}: ARI over rows and columns in function of $\pmis$, the proportion of missing links
        for the four colBiSBM models and their LBM counterparts.
        Each dataset is generated with the corresponding colBiSBM model on a nested structure and
        fitted with the same model and the LBM model labelled as sep to give a baseline of the performance.
    }%
    \label{fig:ari-plot-nested}
\end{figure}

\paragraph{Results} \Cref{fig:auc-plot-nested,fig:ari-plot-nested} respectively present
the results of the simulation for AUC and ARI metrics.

From~\Cref{fig:auc-plot-nested} we see result similar
to~\Cref{fig:auc-plot-modular}, our models always perform at least as best as
the baseline sep and with smaller variability. But the performance
difference is marginal and non significant except in very degraded case
($p_{\text{mis}}\geq 0.7$).

On~\Cref{fig:ari-plot-nested}, we see trends similar
to~\Cref{sssec:on-modular-structure}. All models at worst equal
their sep counterpart (apart from iid for rows at $p_{\text{mis}}
    = 0.6$). For the iid and $\rho$ datasets the sep is having
trouble on columns and rows respectively.

Four main differences from~\Cref{sssec:on-modular-structure}
are that iid-colBiSBM gives poor performance node membership attribution
for columns, for $\rho$-colBiSBM on rows it performs a bit better than the
sep but not exceeding $0.5$, still for $\rho$-colBiSBM on columns the
ARI is at $1$ until a sudden variability increase and drop around
$p_{\text{mis}} \geq 0.8$ and the $\pi\rho$-colBiSBM is less drastically bad
around $0.55$ slowly decreasing to $0.1$ but still beating the sep.

What we learn
from~\Cref{sssec:on-modular-structure,sssec:simulation-on-nested-structure}
is twofold. The structure of connectivity of the networks can drastically
change the performance of the inference algorithm. We see that nested
structures resist to higher levels of degradation than modular structures.
In the end the joint modeling still
benefits from the information transfer it allows even in harder settings.

\subsection{Clustering of simulated networks}\label{ssec:clustering-of-simulated-networks}

\paragraph{Simulation settings} For all models we simulate $M = 30$ networks with~$\forall m \in \{ 1 \dots M
    \} , n^m_1 = n^m_2 = 75$ with $Q_1 = Q_2 = 3$.\newline For the simulations the
proportions are the following:\newline $\bm{\pi}^1 = \left( 0.2, 0.3, 0.5
    \right), ~\bm{\rho}^1 = \left( 0.2, 0.3, 0.5 \right)$ and for all $m =
    2,\dots,9$

\begin{align*}
    \bm{\pi}^m = \begin{cases}
                     \bm{\pi}^1             & \text{for iid}                      \\
                     \sigma_1^m(\bm{\pi}^1) & \text{for } \pi \text{ and } \pi\rho
                 \end{cases},&\qquad
    \bm{\rho}^m =
    \begin{cases}
        \bm{\rho}^1             & \text{for iid}                       \\
        \sigma_2^m(\bm{\rho}^1) & \text{for } \rho \text{ and } \pi\rho
    \end{cases},
\end{align*}
where $\sigma_1^m$ and $\sigma_2^m$ are permutations of \{1, 2, 3\} proper to network $m$ and
$\sigma_1 (\pi)= {(\pi_{\sigma_1 (i)})}_{i=\{1,\dots,3\}}$
and $\sigma_2 (\rho)= {(\rho_{\sigma_2 (i)})}_{i=\{1,\dots,3\}}$.

The networks are divided into 3 sub-collections of 10 networks with
connectivity parameters as follows:
\begin{align*}
    \bm{\alpha}^{as} = .3 + \begin{pmatrix}
                                \epsilon             & - \frac{\epsilon}{2} & - \frac{\epsilon}{2} \\
                                - \frac{\epsilon}{2} & \epsilon             & - \frac{\epsilon}{2} \\
                                - \frac{\epsilon}{2} & - \frac{\epsilon}{2} & \epsilon
                            \end{pmatrix}, &                                                                     &
    \bm{\alpha}^{dis} = .3 + \begin{pmatrix}
                                 - \frac{\epsilon}{2} & \epsilon             & \epsilon             \\
                                 \epsilon             & - \frac{\epsilon}{2} & \epsilon             \\
                                 \epsilon             & \epsilon             & - \frac{\epsilon}{2}
                             \end{pmatrix},
                                                                          & \bm{\alpha}^{cp} = .3 + \begin{pmatrix}
                                                                                                        \frac{3 \epsilon}{2} & \epsilon           & \frac{\epsilon}{2}   \\
                                                                                                        \epsilon             & \frac{\epsilon}{2} & 0                    \\
                                                                                                        \frac{\epsilon}{2}   & 0                  & - \frac{\epsilon}{2}
                                                                                                    \end{pmatrix} &
\end{align*}
with $\epsilon \in [.1, .4]$. $\bm{\alpha}^{as}$ represents a classical
assortative community structure, while $\bm{\alpha}^{cp}$ is a layered
core-periphery structure with block 2 acting as a semi-core. Finally,
$\bm{\alpha}^{dis}$ is a dis-assortative community structure with stronger
connections between blocks than within blocks. If $\epsilon = 0$, the three
matrices are equal and the 9 networks have the same connection structure thus
making the node membership attribution and partitioning of the different
structure types difficult. Increasing $\epsilon$
differentiates the 3 sub-collections of networks and makes the classification
task easier.

\begin{figure}[!t]
    \centering
    \includegraphics[width=0.8\textwidth]{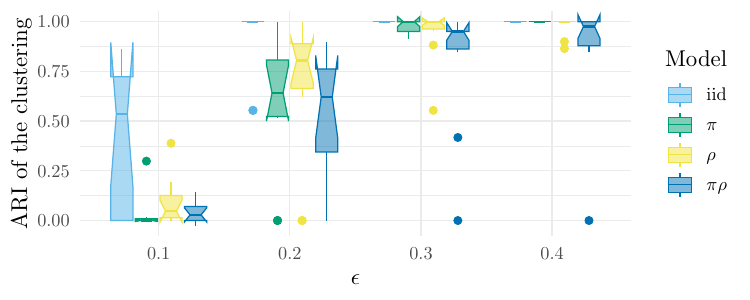}
    \caption{ARI obtained for the partitioning with the different models in
        function of $\epsilon$}
    \label{fig:ari-partitioning-boxplot}
\end{figure}

\begin{figure}
    \centering
    \includegraphics[width=0.8\textwidth]{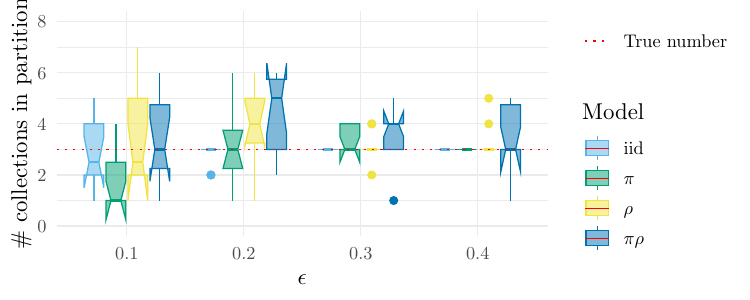}
    \caption{Number of collections in partition obtained for the clustering with the different models in function of $\epsilon$. The true number of collections is 3 plotted by the red dashed line.}\label{fig:nbcollections-clustering}
\end{figure}

\paragraph{Results} The evaluation of our method involves a comparison between the resulting
partition of the network collection and the simulated partition using the ARI
metric (\Cref{fig:ari-partitioning-boxplot}). As the value of $\epsilon$ increases, our ability to distinguish
between the networks improves, and this distinction becomes nearly perfect in
all setups of the colBiSBM as soon as $\epsilon \geq 0.3$.

The number of collections in the partition obtained for the clustering with the
different models is plotted in~\Cref{fig:nbcollections-clustering}. There is a
variation in the number of collections yielded by the algorithm for the iid,
$\pi$ and $\rho$ models the medians settle around true value for $\eps[\alpha]
    \geq 0.3$ while for the $\pi\rho$ it happens for $\eps[\alpha] = 0.4$. This may
be due to the fact that there is more freedom in the $\pi\rho$ model and thus
more variability. Looking at the number of partition helps understand why some partitioning yields an ARI of $0$, this is due to the partitioning algorithm deciding to stop after fitting the full collection and not proposing any cut.

\section{Data for applications on ecological networks}
\label{sec:data-for-sec-applications-on-ecological-networks}
\begin{table}[H]
    \centering
    \caption{Shared species over total species for both pollinators and
        plants for the 5 Baldock networks.\label{tab:baldock-shared-species}}
    \renewcommand{\arraystretch}{1.5}
    \begin{tabular}{llllll}
        \hline
                  & TB+JN                         & Bristol                         & Edinburgh                       & Leeds                           & Reading \\
        \hline
        TB+JN     &                               &                                 &                                 &                                 &         \\
        Bristol   & $\frac{0}{237},\frac{1}{301}$ &                                 &                                 &                                 &         \\
        Edinburgh & $\frac{0}{181},\frac{1}{250}$ & $\frac{73}{257},\frac{73}{236}$ &                                 &                                 &         \\
        Leeds     & $\frac{0}{198},\frac{1}{247}$ & $\frac{76}{271},\frac{79}{227}$ & $\frac{61}{230},\frac{58}{197}$ &                                 &         \\
        Reading   & $\frac{0}{200},\frac{1}{288}$ & $\frac{73}{276},\frac{89}{258}$ & $\frac{66}{227},\frac{62}{234}$ & $\frac{69}{241},\frac{74}{219}$ &         \\
        \hline
    \end{tabular}
    \renewcommand{\arraystretch}{1}
\end{table}

\fi

\end{document}